\documentclass[final,3p,times,twocolumn]{elsarticle}
\usepackage{graphicx}
\usepackage{url}
\usepackage{wrapfig}
\usepackage{algorithmic}
\usepackage{algorithm}
\usepackage{float}
\usepackage{color}

\usepackage{amssymb}

\def \figurewidth {5.2cm}
\def \graphwidth {4.2cm}
\def \tablespacer {0.2em}
\def\aligntop#1{\vtop{\null\hbox{#1}}}

\newcommand{\figref}[1]{Fig.~\ref{#1}}

\newcommand{\algref}[1]{Alg.~\ref{#1}}
\newcommand{\tabref}[1]{Table~\ref{#1}}

\newcommand{\shortcite}[1]{\cite{#1}}
\newcommand{\shortciteetal}[1]{{et. al.}~\shortcite{#1}}

\begin{document}

\begin{frontmatter}



\title{Table-Top Scene Analysis Using Knowledge-Supervised MCMC}


\author{Ziyuan Liu\fnref{label2}} \ead{ziyuan.liu@tum.de} \author{Dong Chen} \ead{chendong@mytum.de}
\address{Siemens AG, Corporate Technology, Munich, Germany\\
Institute of Automatic Control Engineering, Technische Universit\"at M\"unchen, Munich, Germany
 }

\fntext[label2]{Corresponding author. Postal address: Karlstr. 45, room 5001, 80333, Munich, Germany. Telephone: +49-89-289-26900. Fax: +49-89-289-26913.}


\author{Kai M. Wurm}
\address{Siemens AG, Corporate Technology, Munich, Germany}

\author{Georg von Wichert}
\ead{georg.wichert@siemens.com}
\address{Siemens AG, Corporate Technology, Munich, Germany \\ Institute for Advanced Study, Techniche Universit\"at M\"unchen, Lichtenbergstrasse 2a, D-85748 Garching, Germany}

\begin{abstract}

In this paper, we propose a probabilistic approach to
generate abstract scene graphs from uncertain 6D pose estimates. We focus on generating a semantic understanding of the perceived scenes that well explains the composition of the scene and the inter-object relations. The proposed system is realized by our knowledge-supervised MCMC sampling technique. We explicitly make use of task-specific context knowledge by encoding this knowledge as descriptive rules in Markov logic networks. We use a probabilistic sensor model to encode the
fact that measurements are subject to significant uncertainty. We
integrate the measurements with the abstract scene graph in a data
driven MCMC process. Our system is fully probabilistic and links the high-level abstract scene description to uncertain low level measurements. Moreover, false estimates of the object poses and hidden objects of the perceived scenes can be systematically detected using the defined  Markov logic knowledge base. The effectiveness of our approach is demonstrated and evaluated in real world experiments.
\end{abstract}


\end{frontmatter}



\section{Introduction}
\begin{figure*}
	\centering
	\includegraphics[width=1.4\columnwidth]{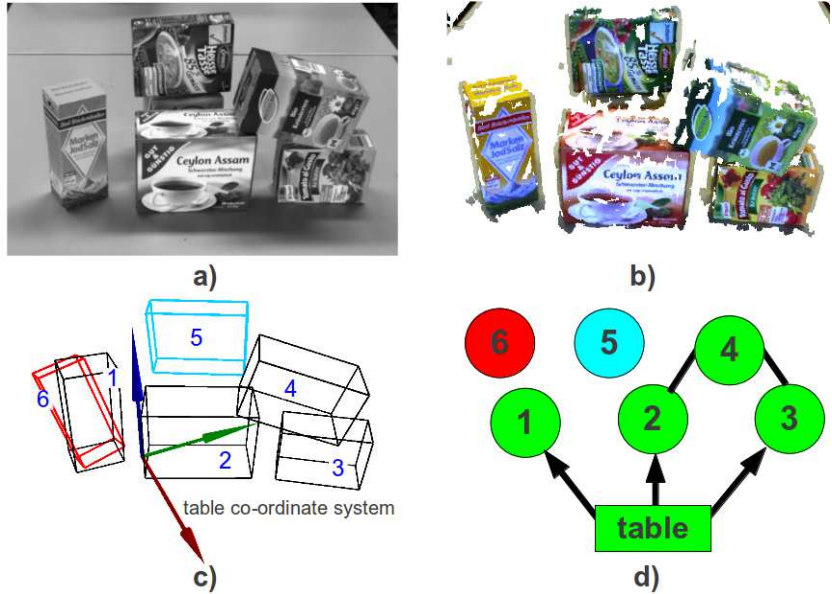}
    \caption{An example output of our system. a) - b) Sensor input for
      6D pose estimation: stereo image (a) or 3D point cloud (b). c)
      Initial guess of object 6D poses obtained by a feature-based
      approach. The three axes of the table coordinate system are
      shown as blue, red and green arrows. d) The scene graph
      generated by our system. Arrows indicate the relation ``stable
      support". Undirected lines show the relation ``unstable
      contact". Object \#5 is considered to have a ``hidden object"
      under it. object \#6 is considered to be a ``false estimate".}
    \label{fig-p3:system overview}
\end{figure*}

For autonomous robots to successfully perform manipulation tasks, such
as cleaning up and moving things, they need a structural understanding
of their environment. It is not sufficient to provide geometry scene
knowledge alone, i.e., the locations of the objects relevant to the
manipulation task. The robots planning components require additional
information about the composition and inter-object relations within
the scene. Imagine a robot that is asked to fetch one of
the objects shown in~\figref{fig-p3:system overview}.  It is important
for the robot to understand for example that
\begin{itemize}
	\item to move object \#3, object \#4 should be moved first, otherwise object \#4 will fall while moving object \#3.
	\item object \#6 is a false estimate thus  can not be moved.
	\item there is something hidden under object \#5.
\end{itemize}

In this paper, we propose a probabilistic method to generate abstract
scene graphs for table-top scenes that can answer such questions. The
input to our algorithm is 6D object poses that are generated using a
feature-based pose estimation approach. Object poses can be estimated
either from stereo images or from RGBD point clouds. A typical result
of the procedure is shown in Fig.~\ref{fig-p3:system overview}.

Our scene graph for table-top scenes describes the composition of the
perceived scene and the relations between the objects, such as
``support" and ``contact". To efficiently generate such scene graphs,
we explicitly formulate and use context knowledge, which we encode in
logic rules that typically hold for table-top scenes, e.g., ``objects
do not hover over the table", or ``objects do not intersect with each
other".

\indent The remainder of this paper is structured as follows: in section \ref{sec:related}, we review related work in the field of context-based scene analysis. In section \ref{sec:contribution} we list our contributions. In section \ref{sec:gkf}, we explain the fundamental idea of our generalizable knowledge framework. In section \ref{sec:mln}, we briefly introduce the theory of Markov logic networks. In section \ref{sec:system}, we elaborate on how to use our knowledge framework to solve the problem of table-top scene analysis. In section \ref{sec:exp}, we evaluate the performance of our system using real world data. In the end,  we conclude in section \ref{sec:sum}.

\section{Related Work}
\label{sec:related}


Scene analysis involves several different aspects, such as
object identification \cite{sun2013attribute,dale1995computational},
object localization \cite{fulkerson2009class, lampert2009efficient,
  harzallah2009combining}, and object discovery
\cite{karpathy2013object,kang2011discovering,
  weber2000towards}. Depending on the context, each individual scene
can be very different. For a table-top scene, a scene may contain
several objects that are commonly found on tables, such as, books and
computers. In traffic analysis, a scene may be something completely
different and consist of cars, traffic lights and other relevant
objects. The goal of different applications of scene analysis is not
the same either. Some approaches concentrate on identifying and
localizing the objects involved, while other approaches try to
discover objects in a cluttered environment. In the following, we
provide a short review on existing approaches, and we focus on the
ones that use context information to analyze the perceived scene.

Several previous methods represent context knowledge as descriptive
logic rules to help robots understand the perceived scenes. Using
description logic
\cite{baader2003description,grosof2003description,calvanese1998description},
ontologies \cite{hofweber2008logic,valore2006topics} are used to
encode knowledge about the composition of scenes, for example, that a
set of cutlery consists of a knife, a fork, and a spoon. Using
reasoning engines, such as Racer \cite{haarslev2003racer}, Pellet
\cite{sirin2006pellet} and FaCT++ \cite{tsarkov2006fact++}, missing or
wrong items in the scene can be inferred so as to give robots
higher-level understanding of the perceived scene. In addition, robot
actions are sometimes also encoded as ontologies. Different steps of
performing a certain task, such as setting up a table, are defined as
the composites. Based on inference results of the defined ontologies,
corresponding actions are triggered for the operating robot.


Lim~\shortciteetal{lim2011ontology} presented an ontology-based knowledge framework, in which they model robot knowledge as a semantic network. This framework is comprised of two parts: knowledge description and knowledge association. Knowledge description combines knowledge regarding perceptual features, part objects, metric maps, and primitive behaviors with knowledge about perceptual concepts, objects, semantic maps, tasks, and contexts. Knowledge association adopts both unidirectional and bidirectional rules to perform logical inference. This framework enabled their robot to complete a ``find a cup" task in spite of hidden or partial data. 

Tenorth~\shortciteetal{tenorth2010knowrob} proposed a system for
building environment models for robots by combining different types of knowledge. Spatial information about objects in the environment is combined with encyclopedic knowledge to inform robots of the types and properties of objects. In addition, common-sense knowledge is used to describe the functionality of the involved objects. Furthermore, by learning statistical relational models, another type of knowledge is derived from observations of human activities. By providing robots deeper knowledge about objects, such as their types and their functionalities, this system helps robots accomplish complex tasks like cleaning dishes. 


Pangercic~\shortciteetal{pangercic2010combining} proposed a top-down guided 3D model-based vision algorithm for assisting household environments. They use ``how-to" instructions which are parsed and extracted from the wikihow.com webpage \cite{wikihow} to shape the top-down guidance. The robots knowledge base is represented in Description Logics (DL) using the Web Ontology Language (OWL) \cite{mcguinness2004owl}. Based on the knowledge base, inferences are obtained using SWI-Prolog queries \cite{wielemaker1996swi}. Using this system, the task ``how to set a table" is accomplished, in which a robot prepares a table for a meal according to the instructions obtained from the wikihow webpage. 


Some other methods use
first-order logic \cite{smullyan1995first, fitting1996first} or
variants of first-order logic, such as Markov logic networks
\cite{richardson2006markov} or Bayesian logic networks
\cite{jain2009bayesian}, to formulate context knowledge to solve scene
analysis. Blodow~\shortciteetal{blodow2010perception} use a Markov
logic network to identify objects. Instead of generating scene graphs
of the perceived scenes, they focus on inferring the temporal
correspondence between observations and entities. By keeping track of
where objects of interest are located, they aim to provide robots
environment awareness, so that robots are able to infer which
observations refer to which entities in the real world.



Additionally, there exist also methods that exploit context information through human robot interaction (HRI) \cite{goodrich2007human}. Motivated from psycholinguistic studies, Swadzba~\shortciteetal{swadzba2009computational} proposed a computational model for arranging objects into a set of dependency trees via spatial relations extracted from human verbal input. Assuming that objects are arranged in a hierarchical manner, they predict intermediate structures which support other object structures, such as, ``soft toys lie on the table". The objects at the leaves of the trees are assumed to be known and used to compute potential planar patches for their parent nodes. The computed patches are adapted to real planar surfaces, so that wrong object assignments are corrected. In addition, new object relations which were not given in the verbal descriptions could also be introduced. In this way, they could generate a model of the scene through context encoded in the verbal input of human observers. Other examples of this kind of approaches can be found in \cite{yu2011active}, \cite{sun2013attribute}, and \cite{mavridis2006grounded}.


A number of approaches analyze scenes using context information in other form. Grundmann~\shortciteetal{grundmann2010probabilistic} proposed a method to increase the estimation accuracy of independent sub-state estimation using statistical dependencies in the prior. The dependencies in the prior are modeled by physical relations. They use a physics engine to test the validity of the sampled physical relations. Scene models that fail the validity check of the implemented physics engine are given a probability of zero. In this way, a better approximation of the joint posterior is achieved. Another example of using physics engines to check scene validity was presented in~\cite{wagle2010multiple}. 

By modeling the relations between objects and their supporting surfaces in the image as a graphical model, Bao~\shortciteetal{bao2012toward} formulate the problem of objects detection as an optimization problem, in which parameters such as the object locations or the focal length of the camera are optimized. They follow the intuition that objects�location and pose in the 3D space are not arbitrarily distributed but rather constrained by the fact that objects must lie on one or multiple supporting surfaces. Such supporting surfaces are modeled by means of hidden parameters. The solution to the problem is finding the set of parameters that maximizes the joint probability. However, this approach aimed at detecting objects using context information and did not provide an abstract understanding of the perceived scenes.


%


\subsection*{Contributions}
\label{sec:contribution}
Other than the approaches introduced above, we propose a probabilistic approach to
generate abstract scene graphs from uncertain 6D pose estimates. We focus on generating a semantic understanding of the perceived scenes that well explains the composition of the scene and the inter-object relations. The proposed system is realized by our knowledge-supervised MCMC sampling technique \cite{liu2013generalizable}. We employ \emph{Markov Logic Networks} (MLNs) \cite{richardson2006markov} to encode the underlying context knowledge, since MLNs are able to model
uncertain knowledge by combining first-order logic with probabilistic
graphical models. We use a probabilistic sensor model to encode the
fact that measurements are subject to significant uncertainty. We
integrate the measurements with the abstract scene graph in a data
driven MCMC process. Our system is fully probabilistic and links the high-level abstract scene description to uncertain low level measurements. Moreover, false estimates of the object poses and hidden objects of the perceived scenes can be systematically detected using the Markov logic inference techniques.

\section{A Generalizable Knowledge-Supervised MCMC Sampling Framework}
\label{sec:gkf}

Markov chain Monte Carlo (MCMC) \cite{neal1993probabilistic} methods are a class of algorithms for sampling from probability distributions. In MCMC methods, a Markov chain, whose equilibrium distribution is identical with the desired distribution, is constructed by sequentially executing state transitions according to a proposal distribution. The state of the chain after certain burn-in time is then used as a sample of the desired distribution.

In this paper, we apply our generalizable knowledge-supervised MCMC
(KSMCMC) sampling framework~\cite{liu2013generalizable}, which is a modern extension of MCMC methods, to interpret
table-top scenes. KSMCMC is a combination of Markov logic networks \cite{richardson2006markov} and data-driven MCMC sampling \cite{tu2005image}. Based on Markov logic networks, task-specific context knowledge can be formulated as descriptive logic rules. These rules define the system behaviour on higher levels and can be processed by modern knowledge reasoning techniques. Using data-driven MCMC, samples can be efficiently drawn from unknown complex distributions. As a whole, KSMCMC is a new method of fitting abstract semantic models to input data by combining high-level knowledge processing with low-level data processing in a probabilistic and systematic way.

The fundamental idea of our framework is to define
an abstract model $M$ to explain data $\textrm{D}$ with the help of
rule-based context knowledge (defined in MLNs). According to Bayes'
theorem, a main criterion for evaluating how well the abstract model
$M$ matches the input data $\textrm{D}$ is the \emph{posterior}
probability of the model conditioned on the data $p({M}|{\textrm{D}})$
which can be calculated as follows:
\begin{equation}
p({M}|\textrm{D})\propto p(\textrm{D}|{M})\cdot p({M}).
\label{eq:pos}
\end{equation}
Here, the term $p(\textrm{D}|{M})$ is usually called the \emph{likelihood}
and indicates how probable the observed data are for different
settings of the model. The term $p({M})$ is the \emph{prior}
describing what kind of models are possible at all. We propose to
realize the prior by making use of context knowledge in the form of
descriptive rules, so that the prior distribution is shaped in such a
way that impossible models are ruled out. Calculations of the prior and likelihood are explained in section \ref{sec:prior-calc} and section \ref{sec:likelihood-calc} respectively.

Starting from an initial guess of the model, we apply a data driven MCMC process to improve the quality of the abstract
model. Our goal is then to find the model ${M}^*$ that best explains
the data and meanwhile complies with the prior, which leads to the
maximum of the posterior probability:
\begin{equation}
{M}^*=\arg\!\max_{\!\!\!\!\!\!\!\!\!\!\!{\small{{M}}}\in\Omega} \, p({M}|\textrm{D}),
\label{eq:argmax}
\end{equation}
where $\Omega$ indicates the entire solution space. Details on this process are provided in section \ref{sec: ddmcmc}.


\section{Markov Logic Networks}
\label{sec:mln}
\indent Before explaining the theory of Markov Logic Networks (MLNs),
we briefly introduce the two fundamental ingredients of MLNs, which
are Markov Networks and First-Order Logic.

\subsection{Markov Networks}
\indent According to \cite{pearl1988probabilistic}, a Markov network is a model for representing the joint distribution of a set of variables $X=(X_1, X_2,\dots,X_n)\in \mathbb{X}$, which constructs an undirected Graph $G$, with each variable represented by a node of the graph. In addition, the model has one potential function $\phi_k$ for each clique in the graph, which is a non-negative real-valued function of the state of that clique. Then the joint distribution represented by a Markov network is calculated as
\begin{equation}
P(X=x)=\frac{1}{Z}\prod_{k}\phi_k(x_{\{k\}}),
\label{eq:markov}
\end{equation}
with $x_{\{k\}}$ representing the state of the variables in the $k$th clique. The partition function $Z$ is calculated as
\begin{equation}
Z=\sum_{x\in\mathbb{X}}\prod_k\phi_k(x_{\{k\}}).
\end{equation}
By replacing each clique potential function with an exponential weighted sum of features of the state, Markov networks are usually used as log-linear models:
\begin{equation}
P(X=x)=\frac{1}{Z}\exp\left(\sum_{j}\omega_jf_j(X)\right),
\label{eq:markov-2}
\end{equation}
where $f_j(x)$ is the feature of the state and it can be any real-valued function. For each possible state $x_{\{k\}}$ of each clique, a feature is needed with its weight $\omega_j=\log\phi_k(x_{\{k\}})$. Note that for the use of MLNs only binary features are adopted, $f_j(x)\in\{0,1\}.$ For more details on Markov networks, please refer to \cite{pearl1988probabilistic}.

\subsection{First-Order Logic}
Here we briefly introduce some definitions in first-order logic, which are needed to understand the concept of Markov logic networks. For more details on first-order logic, please refer to \cite{genesereth1987logical}.
\begin{itemize}
	\item \textit{Constant} symbols: these symbols represent objects of the interest domain.
	\item \textit{Variable} symbols: the value of these symbols are the objects represented by the constant symbols.
	\item \textit{Predicate} symbols: these symbols describe relations or attributes of objects.
	\item \textit{Function} symbols: these symbols map tuples of objects to other objects.
	\item An \textit{atom} or \textit{atomic formula} is a predicate symbol used to represent a tuple of objects.
	\item A \textit{ground atom} is an atom containing no variables.
	\item A \textit{possible world} assigns a truth value to each possible ground atom.
\end{itemize}

Together with logical connectives and quantifiers, a set of logical
formulas can be constructed based on atoms to build a
\textit{first-order knowledge base}.

\subsection{MLNs}
\indent Unlike first-order knowledge bases, which are represented by a
set of hard formulas (constraints), Markov logic networks soften the
underlying constraints, so that violating a formula only makes a world
less probable, but not impossible (the fewer formulas a world
violates, the more probable it is). In MLNs, each formula is assigned
a weight representing how strong this formula is. The definition of a MLN is~\cite{richardson2006markov}:

\textit{
A Markov logic network $L$ is a set of pairs ($F_i,\omega_i$), where $F_i$ is a formula in first-order logic and $\omega_i$ is a real number. Together with a finite set of constants $C=\{c_1,c_2,\dots,c_{|C|}\}$, it defines a Markov network $M_{L,C}$ as follows:
\begin{enumerate}
	\item[1.] $M_{L,C}$ contains one binary node for each possible grounding of each predicate appearing in $L$. The value of the node is 1 if the ground atom is true, and 0 otherwise.
	\item[2.] $M_{L,C}$ contains one feature for each possible grounding of each formula $F_i$ in $L$. The value of this feature is 1 if the ground formula is true, and 0 otherwise. The weight of the feature is the $\omega_i$ associated with $F_i$ in $L$.
\end{enumerate}
}

\indent The probability over possible worlds $x$ specified by the ground Markov network $M_{L,C}$ is calculated as
\begin{eqnarray}
P(X=x)=\frac{1}{Z}\exp\left(\sum_i\omega_i n_i(x)\right)\nonumber\\
=\frac{1}{Z}\prod_i\phi_i(x_{\{i\}})^{n_i(x)},
\label{equ:z}
\end{eqnarray}
where $n_i(x)$ is the number of true groundings of $F_i$ in $x$, $x_{\{i\}}$ is the state (truth values) of the atoms appearing in $F_i$, and $\phi_i(x_{\{i\}})=e^{\omega_i}$. For more details on MLN, please refer to \cite{richardson2006markov}.

\section{Scene Graph Generation}
\label{sec:system}

\subsection{Rule-Based Context Knowledge}
\begin{figure}
	\centering
	\includegraphics[width=\columnwidth]{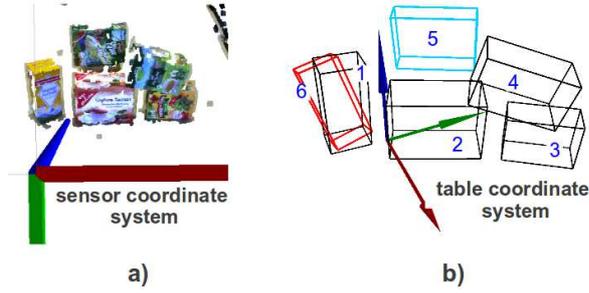}
    \caption{a) The object poses are initially calculated in the
      sensor coordinate system. b) To apply context knowledge in a
      table-top scene, the objects poses are transformed into the
      table coordinate system.}
    \label{fig-p3:table}
\end{figure}

Objects on a table-top are not arranged arbitrarily but they follow
certain physical constraints. In our system, we formulate such
constraints as context knowledge using descriptive rules. This
knowledge helps to model table-top scenes efficiently by ruling out
impossible scenes. We express physical constraints in a table
coordinate system. This table coordinate system can be efficiently
detected from the sensor input, e.g., using the Point Cloud Library
\cite{rusu20113d}.  To apply context knowledge to scene analysis, we
transform the initial guess of the 6D poses of the objects from the
sensor coordinate system into the table coordinate system
(see~Fig. \ref{fig-p3:table}).

\subsubsection{Evidence Predicates}

\begin{table}
\centering
\begin{tabular}{|l|}
\hline
evidence predicates\\\hline\hline
stable(object)\\\hline
table(object)\\\hline
contact(object,object)\\\hline
intersect(object,object)\\\hline
hover(object)\\\hline
higher(object,object)\\
\hline
\end{tabular}
\caption{Declaration of evidence predicates}
\label{TAB:evidence}
\end{table}

Evidences are abstract terms that are detected from the perceived
scene given the object poses and models. To formulate the knowledge as
descriptive rules, we first define several evidence predicates that
describe the properties of table-top scenes. The predicates are
given in~\tabref{TAB:evidence} and encode the following properties:
\begin{itemize}
	\item \emph{stable(object)}: this predicate indicates that an
          object has a stable pose, i.e., it stably lies on a
          horizontal plane. For instance, objects \#1, \#2, \#3 and
          \#5 in~\figref{fig-p3:table}-b have a stable pose. Objects
          \#4 and \#6, in contrast, have an unstable pose.

	\item \emph{table(object)}: this predicate provides the
          possibility to model tables as objects, so that we can use
          it in the reasoning.

	\item \emph{contact(object,object)}: this predicate indicates
          whether two objects have contact with each other. In the
          scene shown in Fig. \ref{fig-p3:table}-b, for instance,
          there is contact between object \#2 and \#3, and between
          object \#3 and \#4. By contrast, there is no contact
          between object \#4 and \#5, or between object \#2 and \#5.

	\item \emph{intersect(object,object)}: this predicate
          indicates whether two objects intersect each other. In the
          scene shown in Fig. \ref{fig-p3:table}-b, intersection only
          occurs between object \#1 and \#6. The predicates
          \emph{intersect(object,object)} and
          \emph{contact(object,object)} are mutually exclusive.

	\item \emph{higher(object,object)}: this predicate expresses
          that the position of the first attribute is higher than that
          of the second attribute in the table coordinate system. In
          the scene shown in Fig. \ref{fig-p3:table}-b, for instance,
          this predicate is true for (\#5,\#2), (\#4,\#2) and
          (\#4,\#3).

	\item \emph{hover(object)}: this predicate means that an
          object does not have any contact with other objects
          including the table. In the scene shown in
          Fig. \ref{fig-p3:table}-b, this predicate is only true for
          object \#5.
\end{itemize}

\subsubsection{Context Knowledge Defined as Logic Rules}

\begin{table*}
\centering
\begin{tabular}{|l|ll|}
\hline index $i$&weight $\omega_i$&formula $F_i$ \\ \hline
$r_1$&$\infty$&!higher(o1,o1)\\
$r_2$&$\infty$&!intersect(o1,o1)\\
$r_3$&$\infty$&!contact(o1,o1)\\
$r_4$&$\infty$&contact(o1,o2) $\to$ contact(o2,o1)\\
$r_5$&$\infty$&intersect(o1,o2) $\to$ intersect(o2,o1)\\
$r_6$&$\infty$&higher(o1,o2) $\to$ !higher(o2,o1)\\
$r_7$&$\infty$&table(o1) $\to$ !false(o1)\\
$r_8$&$\infty$&table(o1) $\to$ !hidden(o1)\\
$r_9$&$\infty$&table(o1) $\to$ stable(o1)\\
$r_{10}$&$\infty$&stable(o1) $\land$ stable(o2) $\land$ contact(o1,o2) $\land$ \\
&&higher(o1,o2) $\to$ supportive(o2) $\land$ supported(o1)\\
$r_{11}$&log(0.70/0.30)&supported(o1) $\to$ !hidden(o1) \\
$r_{12}$&log(0.90/0.10)&!stable(o1) $\to$ !supportive(o1)\\
$r_{13}$&log(0.90/0.10)&hover(o1) $\to$ false(o1) v hidden(o1)\\
$r_{14}$&log(0.90/0.10)&intersect(o1,o2) $\to$ false(o1) v false(o2)\\
$r_{15}$&log(0.70/0.30)&supportive(o1) $\to$ !false(o1) \\
$r_{16}$&log(0.90/0.10)&stable(o1) $\to$ !false(o1)\\
\hline
\end{tabular}
\caption{Declaration of rules}
\label{TAB:formulas}
\end{table*}

Having defined the evidence predicates, we formulate context knowledge
as descriptive rules using Markov logic. Knowledge can be defined as
soft rules or hard rules in Markov logic to express
uncertainty. Knowledge that holds in all cases are defined as hard
rules. Hard rules are assigned a weight of $\infty$ in Markov
logic. By contrast, soft rules are used to encode uncertain knowledge
and are given a probabilistic weight in Markov logic representing the
uncertainty of the corresponding knowledge. Here, we define several
hard and soft rules to model table-top scenes
(see~\tabref{TAB:formulas}).

\paragraph{Hard Rules}
\begin{enumerate}
	\item[$r_1$:] This rule expresses the fact that an object cannot be higher than itself.
	\item[$r_2$:] This rule indicates that an object does not intersect with itself.
	\item[$r_3$:] This rule encodes the fact that an object does not have contact with itself.
	\item[$r_4$:] This rule means that the predicate \emph{contact(object,object)} is commutative, i.e., given that object $o_1$ has contact with $o_2$, the statement that object $o_2$ has contact with $o_1$ is true.
	\item[$r_5$:] This rule means that the predicate \emph{intersect(object,object)} is commutative, i.e., given that object $o_1$ intersects with $o_2$, the statement that object $o_2$ intersects with $o_1$ is true.
	\item[$r_6$:] This rule means that the predicate \emph{higher(object,object)} is not commutative, i.e., given that object $o_1$ is higher than $o_2$, the statement that object $o_2$ is higher than $o_1$ is wrong.
	\item[$r_7$:] This rule expresses that in a table-top scene, the table (as an object) is not a false estimate.
	\item[$r_8$:] This rule expresses that in a table-top scene, the table (as an object) is the lowest object in the scene and has no hidden object under it.
	\item[$r_9$:] This rule expresses that in a table-top scene, the table (as an object) has a stable pose.
	\item[$r_{10}$:] This rule describes ``supportive" and ``supported" relations between two objects with a stable pose. These relations do not apply for objects with unstable poses. In the scene shown in Fig. \ref{fig-p3:table}-b, for example, this relation holds between the table and objects \#1, \#2, and \#3 respectively.
\end{enumerate}

\paragraph{Soft Rules}
\begin{enumerate}
 \setcounter{enumi}{10}
	\item[$r_{11}$:] This rule encodes the assumption that an object that is already known to be supported (through rule \#10) is not likely to have a hidden object under it.
	\item[$r_{12}$:] This rule expresses the assumption that an object with an unstable pose is unlikely to be supportive.
	\item[$r_{13}$:] This rule states the assumption that a hovering object is either a false estimate or has a hidden support under it.
	\item[$r_{14}$:] This rule states the assumption that if two objects intersect, then one of them is probably a false estimate.
	\item[$r_{15}$:] This rule indicates the assumption that a supportive object
is unlikely to be a false estimate.
	\item[$r_{16}$:] This rule indicates the assumption that an object with a stable pose is unlikely to be a false estimate.
\end{enumerate}

The choice of the rules is a problem-oriented engineering step, and the rules given here serve as an example of how to encode the properties of typical table-top scenes.


\subsubsection{Query Predicates}

\begin{table}
\centering
\begin{tabular}{|l|}
\hline
query predicates\\\hline\hline
supportive(object)\\\hline
supported(object)\\\hline
hidden(object)\\\hline
false(object)\\
\hline
\end{tabular}
\caption{Declaration of query predicates}
\label{TAB:query}
\end{table}

Using these rules, query predicates are inferred given the
evidence. In principle, the query predicates represent the questions
that Markov logic can answer given the defined knowledge base. These
query predicates are listed in Table \ref{TAB:query} and have the
following interpretations:

\begin{itemize}

  	\item \emph{supportive(object)}: this predicate indicates that the object represented by the attribute physically supports other objects.
  	\item \emph{supported(object)}: this predicate indicates that the object represented by the attribute is physically supported by other objects. \emph{supportive(object)} and \emph{supported(object)} are two auxiliary query predicates which are used to infer about \emph{hidden(object)} and \emph{false(object)}.
  	\item \emph{hidden(object)}: this predicate expresses that there is an hidden object in the scene under the object that is represented by the
attribute. In the scene shown in Fig. \ref{fig-p3:table}-b, this predicate is true for object \#5.
  	\item \emph{false(object)}: this predicate indicates that the object represented by the attribute is a false estimate. In the scene shown in Fig. \ref{fig-p3:table}-b, this predicate is true for object \#6.
\end{itemize}

\subsubsection{Weights in Log-odd Form}

Rules \#11 to \#16 are soft and are therefore given a weight in the
log-odd form describing our belief on how often the corresponding
uncertain knowledge holds. A weight in the log-odd form $log(p1/p2)$
with $p1, p2\in(0,1)$ and $p1+p2=1$, means that the corresponding rule
holds with the probability of $p1$ \cite{richardson2006markov}. These
weights can either be learned~\cite{lowd2007efficient, huynh2008discriminative, huynh2009max} or manually
designed~\cite{jainknowledge}. In our work, we use two belief levels
$log(0.90/0.10)$ (very sure) and $log(0.70/0.30)$ (relatively sure)
to encode the uncertainty of knowledge. Using Markov logic
inference, we can answer the queries \emph{hidden(object)} and
\emph{false(object)} in the form of a probability.

\subsubsection{Evidence Generation}
\label{subsubsection:evidence generation}

To do inference in MLNs, necessary evidences must be given as
input. In this work, we focus on objects with a regular shape, in
particular, objects that can be well represented by an oriented
bounding box (OBB) \cite{bender2003computergrafik}. However, the aforementioned principles generalize over objects with other shapes, as long as evidences are provided accordingly. In the following we elaborate on how to generate evidences by analyzing the oriented bounding boxes of
detected objects:
\begin{itemize}
	\item $stable(object)$: if any edge of an object OBB is parallel to the vertical axis of the table coordinate system, we define this object to
have a stable pose, i.e., $stable(object)$=True. Examples are shown in Fig. \ref{fig-p3:stable}. Here object \#0, \#1, \#3 and \#4 have a stable pose. In contrast, object \#2 has a unstable pose.

\begin{figure}
	\centering
	\includegraphics[width=.5\columnwidth]{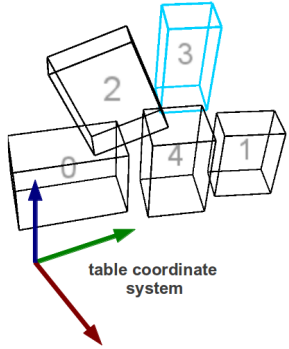}
    \caption{An example scene consisting of five objects represented by their corresponding oriented bounding box.}
    \label{fig-p3:stable}
\end{figure}

	\item \emph{contact(object,object)}: to detect whether two objects have contact with each other, we search for points of intersection between the OBB of these two objects. If two OBBs contact but do not intersect each other, there are three possible cases:
		\begin{itemize}
		\item There is only one point of intersection, and it coincides 			with one of the six vertices of either OBB.
		\item There are multiple points of intersection, and all points are
  		co-linear and lie on one of the twelve edges of either OBB (for
  		example, the contact between object \#2 and \#4
 		in~\figref{fig-p3:stable}).
		\item There are multiple points of intersection, and all points are
  		coplanar and lie on one of the six facets of either OBB (for
  		example, the contact between object \#0 and the table
 	 	in~\figref{fig-p3:stable}).
		\end{itemize}
	In each of the above three cases, we set
\emph{contact(object,object)}=True and
\emph{intersect(object,object)}=False.

	\item \emph{intersect(object,object)}: \emph{contact(object,object)} and \emph{intersect(object,object)} are mutually exclusive, i.e., they can not be true at the same time. If there exist points of intersection between two OBBs, and none of the above cases applies, or if an OBB completely contains the other OBB, then we set \emph{intersect(object,object)}=True. In all other cases, we set \emph{contact(object,object)}=False and \emph{intersect(object,object)}=False. An example of the case that two objects intersect with each other is depicted by Fig. \ref{fig-p3:intersect}. Here \emph{intersect(object,object)} is true for object \#3 and \#5. Points of intersection are shown by gray spheres.

\begin{figure}
	\centering
	\includegraphics[width=.5\columnwidth]{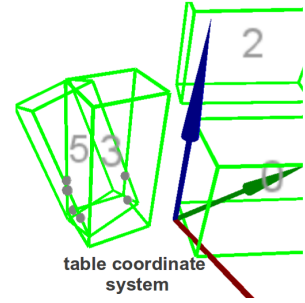}
    \caption{An example of two objects intersecting each other. Points of intersection are shown by gray spheres.}
    \label{fig-p3:intersect}
\end{figure}

	\item \emph{hover(object)}: if an object does not have any
contact or intersection with other objects including the table, then we set \emph{hover(object)}=True. An example for this case is the object \#3 in Fig. \ref{fig-p3:stable}.

	\item \emph{higher(object,object)}: if the position of \emph{object1} is higher than the position of \emph{object2} in the table coordinate system, then we set \emph{higher(object1,object2)}=True.

\end{itemize}

The performance of discriminative generation of evidences is important for our system to deliver correct inferences. If something goes wrong with evidence generation, e.g., a ``supportive" or ``supported" relation is missed, the inference results would be less accurate. Since evidence generation is done based on discriminative methods, errors could happen (but very rarely), when it comes to some near-to-threshold cases. For instance, if we define that two objects are considered to have a contact if the closest distance between these two objects is less than 0.5 cm. Then for the cases, in which the closest distance between two objects is 0.6 cm or 0.7 cm, no contact will be detected, although the detection of contact would be more favourable in this case.


\subsection{Estimation Of Object Poses}
\label{sec:object-localization}
To determine 6D object poses, we apply a pose estimation approach that
is similar to the approach presented by
Grundmann~\shortciteetal{grundmann2010robust}. The basic computational
steps are given in~\algref{alg:object-localization}.  The algorithm is
based on Scale-invariant feature transform (SIFT) keypoints \cite{lowe1999object} that are extracted from triangulated stereo
images or RGBD measurements, e.g., from the Kinect sensor.

\begin{algorithm}
\caption{6D Object Pose Estimation}
\label{alg:object-localization}
\small
\begin{algorithmic}[1]
\REQUIRE {\qquad \\
  {$z$}, input measurement\\
  {$D$}, object database\\
}
\ENSURE {\qquad\\
{$H$}, set of pose hypotheses}\\
\STATE extract SIFT keypoints from $z$
\STATE match keypoints to database $D$
\FOR{all object models $d \in D$}
  \FOR{$i$ iterations}
    \STATE randomly choose three keypoints matched to $d$
    \STATE compute object pose hypothesis from matches
  \ENDFOR
  \STATE cluster pose hypotheses for object $d$
  \STATE add clustered hypotheses to $H$
\ENDFOR
\end{algorithmic}
\end{algorithm}

\begin{figure}
	\centering
	\includegraphics[width=.8\columnwidth]{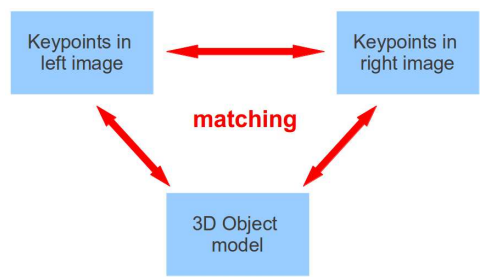}
    \caption{Matching of SIFT keypoints extracted from images.}
    \label{fig-p3:keypoint-matching}
\end{figure}

In a first step, the SIFT keypoints of the observed objects are
matched to a database $D$ of object models.  For each object model $d
\in D$, a maximum of $i$ hypotheses is generated. To generate
hypotheses, three keypoints are chosen randomly from the set of
keypoints that has been matched to model $d$. Here, the keypoints
extracted from the stereo images must undergo a certain matching
scheme to check their validity. The matching scheme is depicted in
Fig. \ref{fig-p3:keypoint-matching}. First of all, the extracted
keypoints are checked by stereo matching, i.e., to check whether a
keypoint found in the left image can also be found in the right image,
or vice versa. The keypoints that have survived the stereo matching
are matched with the object database separately. If a keypoint in the
left image and its corresponding keypoint in the right image (that has
been matched through stereo matching) refer to the same point in the
object database, then this keypoint is a valid keypoint and can be
used for pose estimation.

An object pose hypothesis is then computed from triples of these
matched points. Finally, pose hypotheses are clustered, and outliers
are removed using the RANSAC algorithm~\cite{fischler1981ransac}.

\begin{figure}
	\centering
	\includegraphics[width=.9\columnwidth]{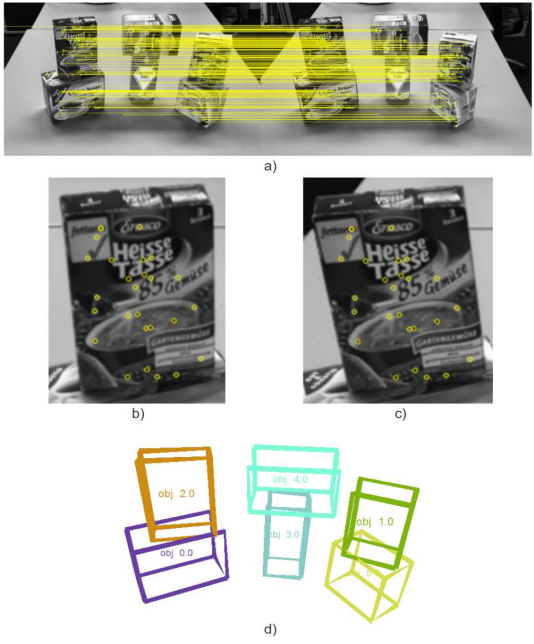}
    \caption{An example of pose estimation. a) Stereo matching of
      detected SIFT keypoints. b) Database-matched keypoints of an object in the left image. c) Database-matched keypoints of the same object in the right image. d) Generated pose hypotheses.}
    \label{fig-p3:pose-estimation}
\end{figure}

An example of pose estimation process is depicted in
Fig. \ref{fig-p3:pose-estimation}. First, the keypoints extracted from
the stereo images are matched
(see~Fig. \ref{fig-p3:pose-estimation}-a, matched keypoint pairs are
visualized by yellow lines). Then, matched keypoints are compared to
the object database. In Fig. \ref{fig-p3:pose-estimation}-b and c,
keypoints found in the object database are shown in yellow. For clarity, only the key points of an object are shown. Using these
matched keypoints, pose hypotheses are generated which are shown in
Fig. \ref{fig-p3:pose-estimation}-d. Pose estimation is performed for
each new scene but is not repeated during scene graph generation.

\subsection{Object Database}

\begin{figure}
	\centering
	\includegraphics[width=.75\columnwidth]{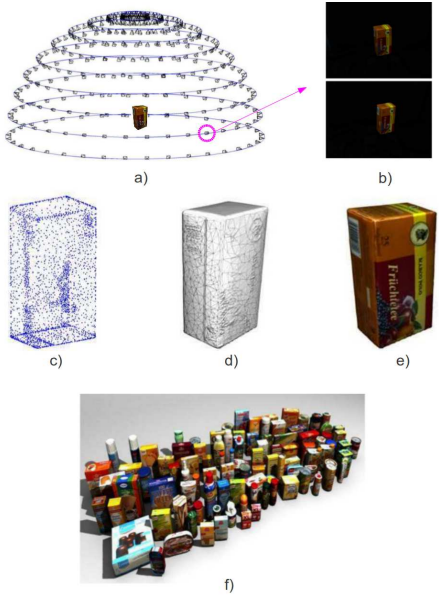}
    \caption{An example of the modelling process. All sub-figures are obtained from \cite{thilo-phd}. a) Camera poses made possible by the turn table and the camera movement. b) A stereo image pair obtained from the highlighted (magenta) camera pose. c) High resolution point cloud obtained through triangulation of matched feature points in stereo images. d) Generated triangle mesh. e) Textured triangle mesh. f) An overview of the object database.}
    \label{fig-p3:3d-model}
\end{figure}

We use the object database of the Deutsche Servicerobotik Initiative
(DESIRE) project \cite{desire}. The object models are generated using an
accurate 3D modelling device which is equipped with a turn table, a
movable stereo camera pair and a digitizer. An example of the modelling process is
illustrated in Fig. \ref{fig-p3:3d-model}. As shown in
Fig. \ref{fig-p3:3d-model}-a, the stereo camera pair first acquires
stereo images of the target object from all possible view angels. Then
2D SIFT keypoints are extracted from these stereo images. Through
keypoints matching and triangulation, a 3D point cloud
(\ref{fig-p3:3d-model}-c) is generated out of the matched 2D SIFT
keypoints. Based on this point cloud and some further optimization
steps, the final object model is generated in the form of a textured
mesh of 3D SIFT keypoints (\ref{fig-p3:3d-model}-e). In
\ref{fig-p3:3d-model}-f, an overview of the DESIRE object database
which contains 100 house-hold items is demonstrated.

\subsection{Calculation of Prior Probability}
\label{sec:prior-calc}
Having defined the predicates and the rules, a knowledge base is formulated in the form of a Markov logic network (MLN). A MLN initializes a ground Markov network \cite{richardson2006markov}, if it is provided with a finite set of constants. In our application, the detected objects and the table form the set of constants. The probability of a possible world $x$ (a hypothesis of scene graph $M$) is given by the probability distribution that is represented by this ground Markov network. As shown in the equation (\ref{equ:z}), this probability is calculated as follows:

\begin{eqnarray}
P(X=x)=\frac{1}{Z}\exp\left(\sum_i\omega_i n_i(x)\right)\nonumber,
\end{eqnarray}
where $n_i(x)$ is the number of true groundings of formula $F_i$ in $x$, and $\omega_i$ is the weight of $F_i$. $Z$ is a normalization factor. As can be seen in the above equation, the probability of a possible world is equal to the exponentiated sum of weights of formulas that are satisfied in this possible world divided by the normalization factor $Z$.
 
By ignoring the normalization factor $Z$, which is the same for all possible worlds, the unnormalized
probability is used as the prior probability in equation (\ref{eq:pos}):
\begin{equation}
p(M)=\exp\left(\sum_i\omega_i n_i(x)\right).
\end{equation}
In this work, we adapt the ProbCog Toolbox \cite{probcog} to perform MLN inference and to calculate this unnormalized probability.

\subsection{Calculation of Likelihood}
\label{sec:likelihood-calc}

To evaluate estimated object poses, we use a Gaussian sensor model as likelihood, which is
similar to the approach proposed by Grundmann~\shortciteetal{grundmann2011gaussian}.  For a pose estimate $\psi$, which corresponds to a scene graph $M$, we first determine the set of keypoints
that have been matched in the object database.  Let
$(x_i,y_i),~i=1,2,\cdots,n$, be the set of 2D image coordinates of the
key points in the stereo image that are matched to the object
database. Using the pin hole camera model \cite{rayleigh1891x},
we project the model keypoints $(x_i,y_i)$ into the image and denote
the resulting set of coordinates as $(x^\psi_i,y^\psi_i)$. The likelihood $p(\textrm{D}|{M})$ in equation (\ref{eq:pos}) is then calculated as
%
%
\begin{equation}
p(\textrm{D}|{M})=\prod^n_i \left( \frac{1}{\sigma_x \sqrt{2\pi}} \; e^{{-\frac{(x_i-x^\psi_i)^2}{2\sigma_x^2}}}
 \frac{1}{\sigma_y \sqrt{2\pi}} \; e^{-\frac{(y_i-y^\psi_i)^2}{2\sigma_y^2}} \right),\label{eq:sensor model}
\end{equation}%
where $\sigma_x$ and $\sigma_y$ are the standard deviation in x- and
y-direction of the image coordinates. We use this sensor model to
determine the likelihood in equation (\ref{eq:pos}). In our
experiments, we use a standard deviation of 1 pixel for $\sigma_x$ and
$\sigma_y$.

\begin{figure}
	\centering
        \begin{tabular}{l l l l}
	  \aligntop{a)} & \aligntop{\includegraphics[height=3.cm]{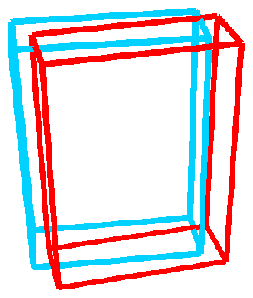}} &
          \aligntop{b)} & \aligntop{\includegraphics[height=3.cm]{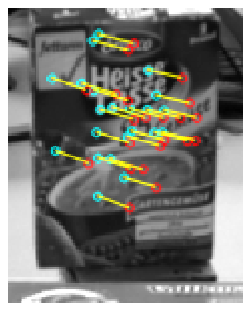}}\\[0.5em]
        \end{tabular}
    \caption{Evaluation of a pose estimate using the Gaussian sensor
      model. a) A pose (red) is evaluated against the database object
      pose (blue). b) Key points in the stereo image that are matched
      with the object model are shown in cyan (for clarity, only the
      left camera image is shown). The projected model key points are
      shown in red. The correspondences between projected and matched
      key points are shown by yellow lines.}
    \label{fig-p3:keypoints-matching}
\end{figure}

An illustration of this method is given in
Fig. \ref{fig-p3:keypoints-matching}. Here, a pose hypothesis (shown in
red) is evaluated against the database object pose (shown in
blue). Key points in the stereo image that are matched to the object
model are shown in cyan. For clarity, only the left camera image is
shown. The projected model key points are shown in red. The
correspondences between projected and matched key points are shown by
yellow lines. The sensor model is calculated based on such
correspondences.

\subsection{Data Driven MCMC}
\label{sec: ddmcmc}
Because sensor data could be noisy and the used object database is imperfect, the input pose estimate of the observed objects is also imperfect. To find the scene graph that best explains the perceived scene, we apply a data driven MCMC process~\cite{tu2005image}. In the
$t$-th iteration with scene graph $M_{t}$, we generate $n$ new pose
estimates $e_{t,i},i=1,2,\cdots,n$, by adding Gaussian noises to the
current pose estimate $e_{t,0}$ and weight them using the sensor model
(equation (\ref{eq:sensor model})). In this way, the input pose estimate is optimized.

\begin{figure}
	\centering
	\includegraphics[width=.5\columnwidth]{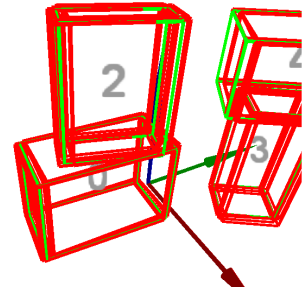}
    \caption{Generating new pose estimates (red) by adding Gaussian noises to the current pose estimate (green).}
    \label{fig-p3:gaussian}
\end{figure}

An example of generating new pose estimates is given in
Fig. \ref{fig-p3:gaussian}. The pose estimate with the best weight
$e_{t}^*$ is used to generate a new scene graph $M_{t+1}$. This
scene graph is accepted by the probability $p_{a}$, using the Metropolis-Hastings
algorithm \cite{chib1995understanding}:

\begin{equation}
p_{a}=\min\left( 1, \frac{P(M_{t+1}|D) \cdot Q(M_{t}|M_{t+1})}{P(M_{t}|D)\cdot Q(M_{t+1}|M_{t})} \right),
\label{icra2014-eq:MH}
\end{equation}
where $P(M_t|D)$ is the posterior probability of $M_t$ (equation (\ref{eq:pos})). $Q(M_{t+1}|M_{t})$ is the proposal probability of generating $M_{t+1}$ out of $M_{t}$ and is calculated as
\begin{equation}
Q(M_{t+1}|M_{t})=\frac{weight(e_{t}^*)}{\sum_i^n weight(e_{t,i})+weight(e_{t,0})}.
\end{equation}
Similarly, $Q(M_{t}|M_{t+1})$ is computed as
\begin{equation}
Q(M_{t}|M_{t+1})=\frac{weight(e_{t,0})}{\sum_i^n weight(e_{t,i})+weight(e_{t,0})}.
\end{equation}

Here $weight(e_{t,i})$ is the likelihood calculated using $e_{t,i}$ as pose estimate (equation (\ref{eq:sensor model})). 

\section{Evaluation}
\label{sec:exp}

We conducted numerous real world experiments to evaluate our
approach. In each experiment, a number of household objects was placed
on a table and a sensor measurement was taken. We then applied our
approach to generate a scene graph and to infer hidden objects or
false estimates.

\begin{figure*}
\centering
\begin{tabular}{p{\figurewidth} p{\tablespacer} p{\figurewidth}}
\includegraphics[width=\figurewidth]{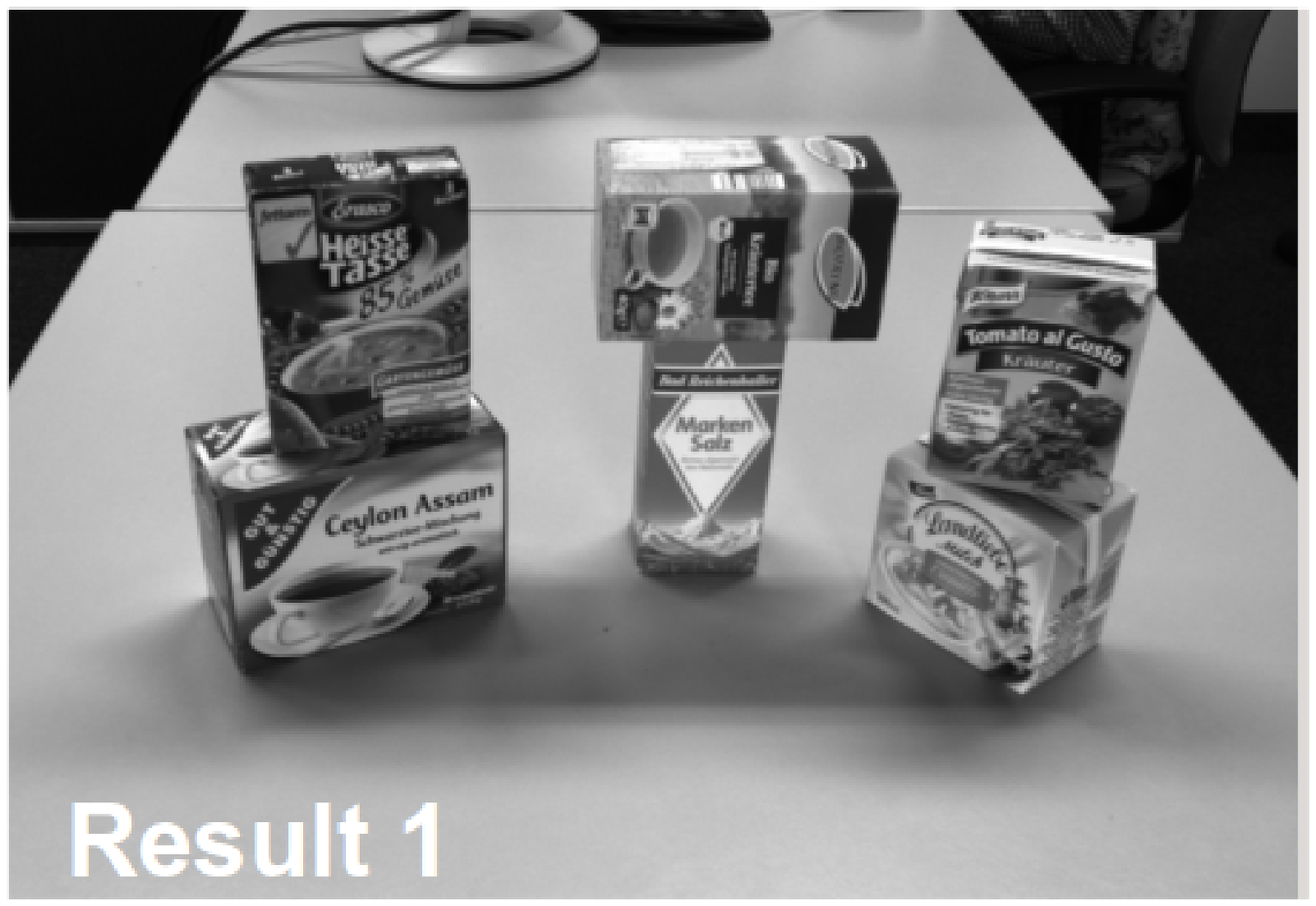} & &\includegraphics[width=\figurewidth]{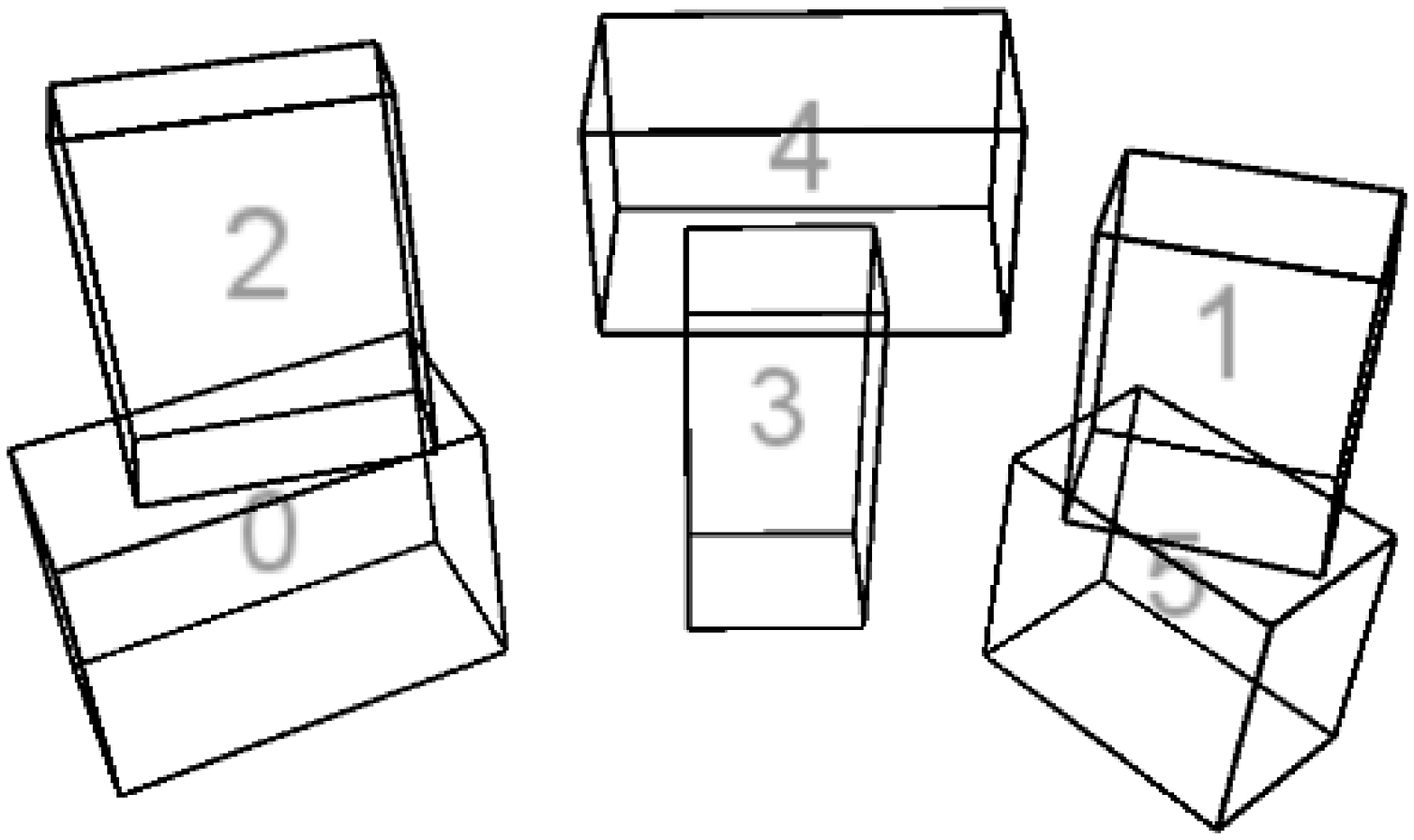}\\ \\
\includegraphics[width=\graphwidth]{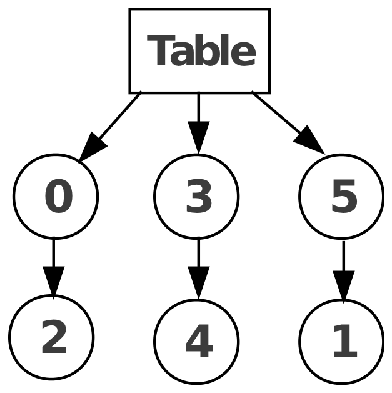}&
\begin{tabular}[b]{ll}
false(0)=0.046, hidden(0)=0.424\\
false(1)=0.070, hidden(1)=0.388\\
false(2)=0.092, hidden(2)=0.392\\
false(3)=0.096, hidden(3)=0.398\\
false(4)=0.142, hidden(4)=0.362\\
false(5)=0.050, hidden(5)=0.422
\end{tabular}
\\[1em]
\end{tabular}
    \caption{Experimental result 1. The input stereo image
      (upper left), estimated 6D poses (upper right), the
      resulting scene graph (lower left) and the query probability (lower right) are shown. False estimates and objects implying hidden objects are highlighted in red and cyan respectively. }
    \label{fig-p3:result1}
\end{figure*}

\begin{figure*}
\centering
\begin{tabular}{p{\figurewidth} p{\tablespacer} p{\figurewidth}}
\includegraphics[width=\figurewidth]{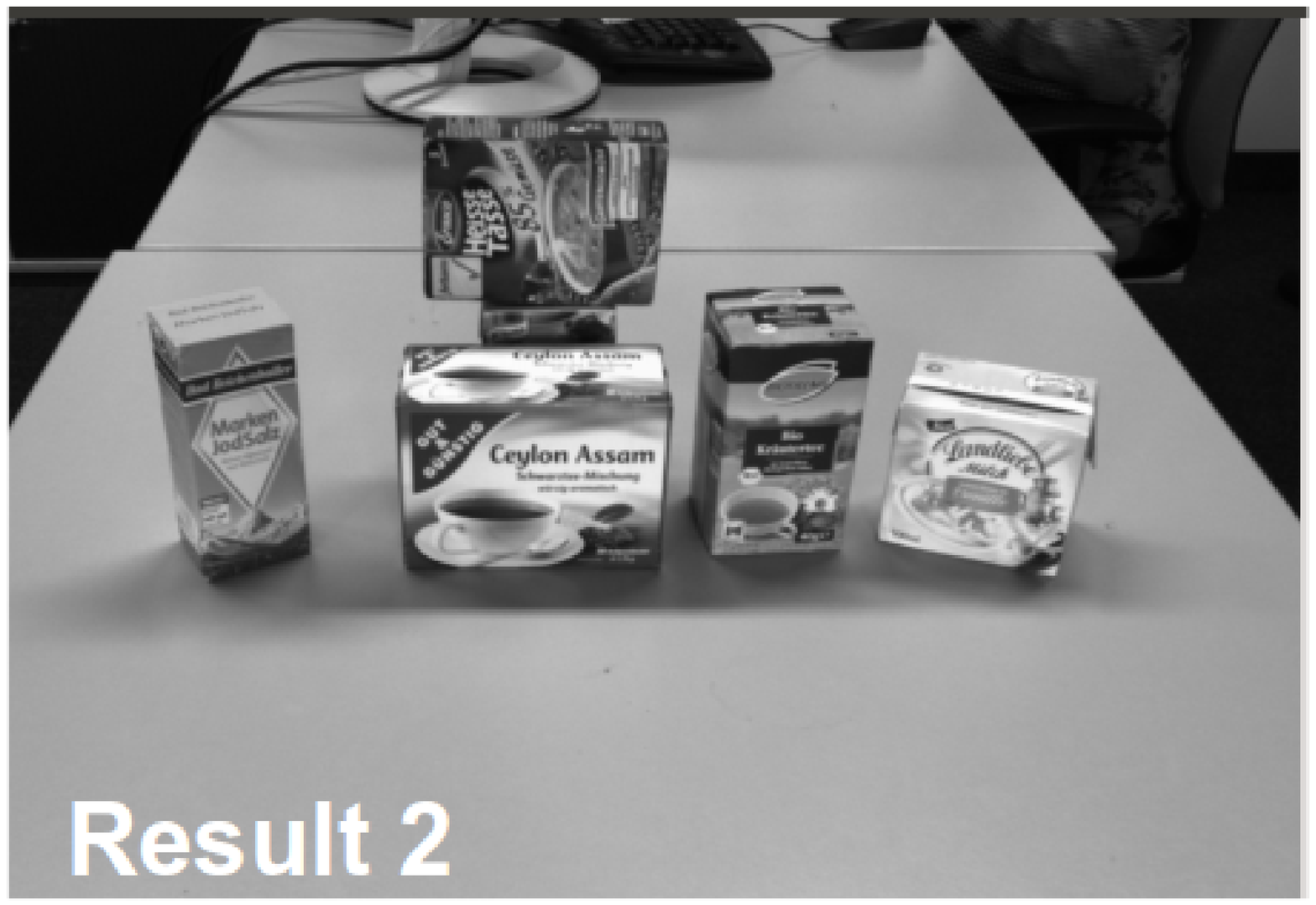} & &\includegraphics[width=\figurewidth]{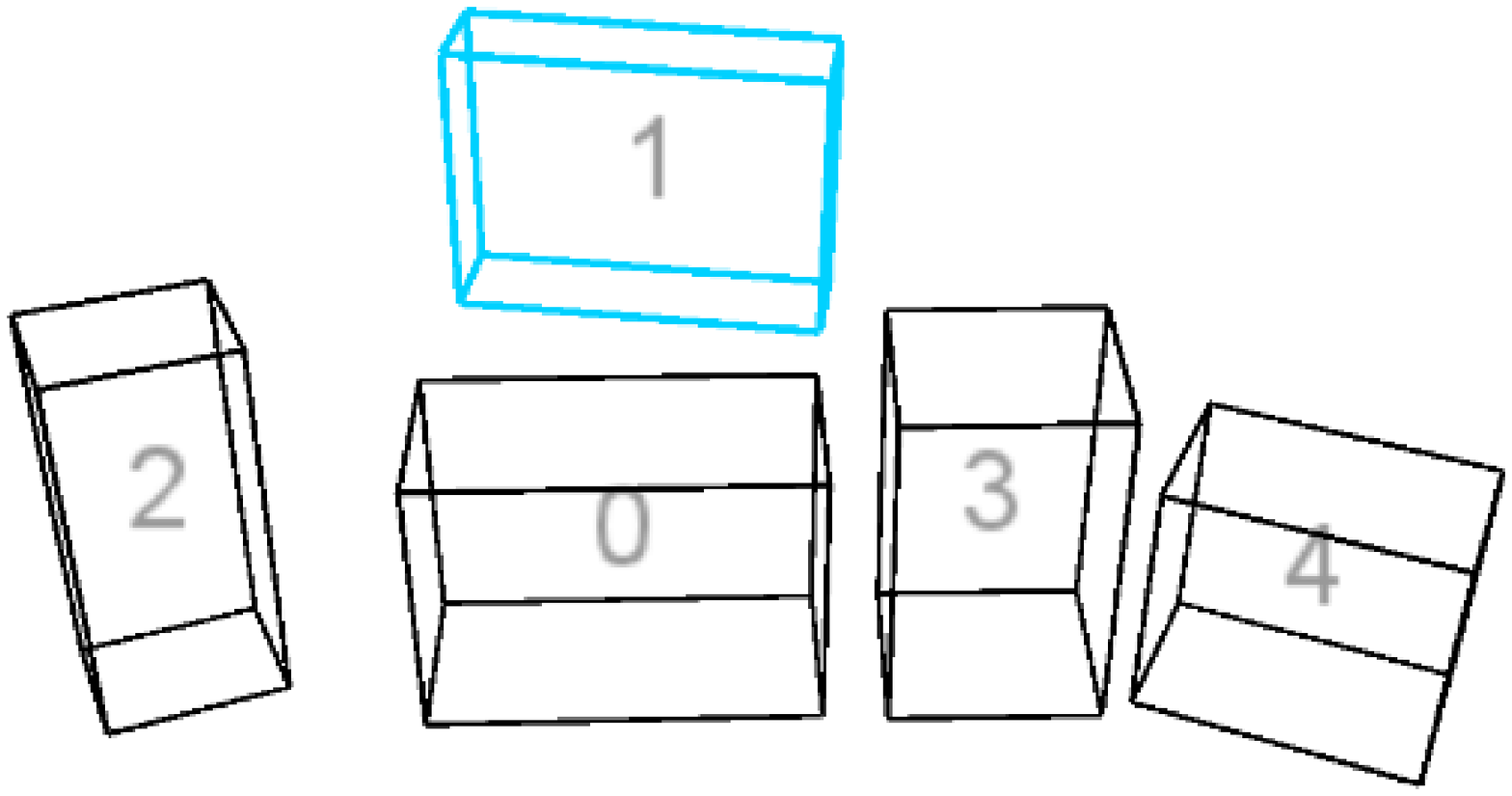}\\ \\
\includegraphics[width=\graphwidth]{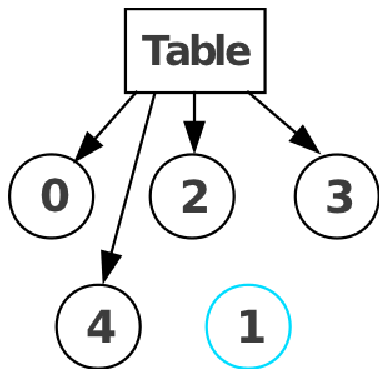}&
\begin{tabular}[b]{ll}
false(0)=0.110, hidden(0)=0.354\\
false(1)=0.126, \textcolor{cyan}{hidden(1)=0.794}\\
false(2)=0.120, hidden(2)=0.384\\
false(3)=0.148, hidden(3)=0.402\\
false(4)=0.094, hidden(4)=0.410
\end{tabular}
\\[1em]
\end{tabular}
    \caption{Experimental result 2. The input stereo image
      (upper left), estimated 6D poses (upper right), the
      resulting scene graph (lower left) and the query probability (lower right) are shown. False estimates and objects implying hidden objects are highlighted in red and cyan respectively. }
    \label{fig-p3:result2}
\end{figure*}

\begin{figure*}
\centering
\begin{tabular}{p{\figurewidth} p{\tablespacer} p{\figurewidth}}
\includegraphics[width=\figurewidth]{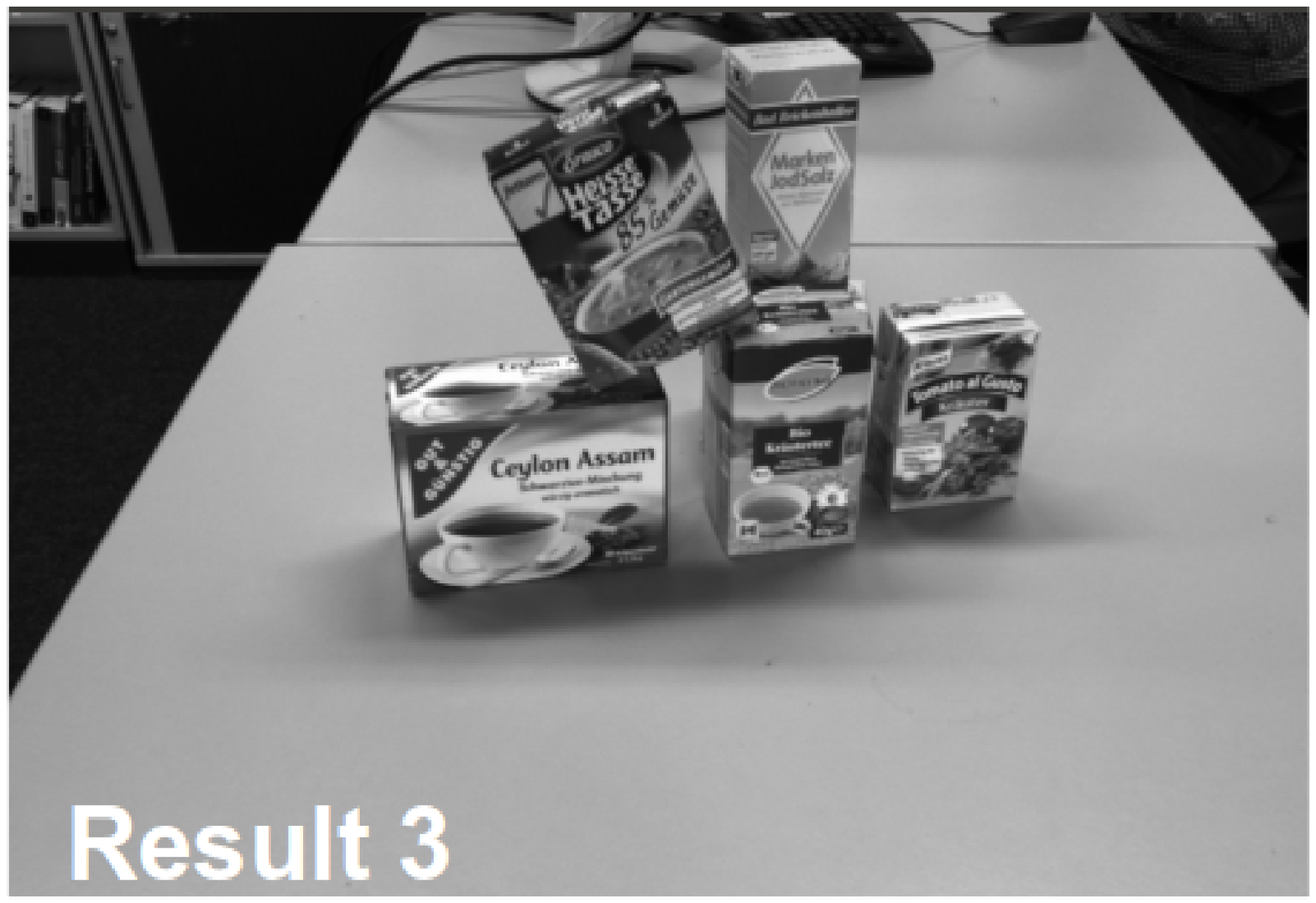} & &\includegraphics[width=\figurewidth]{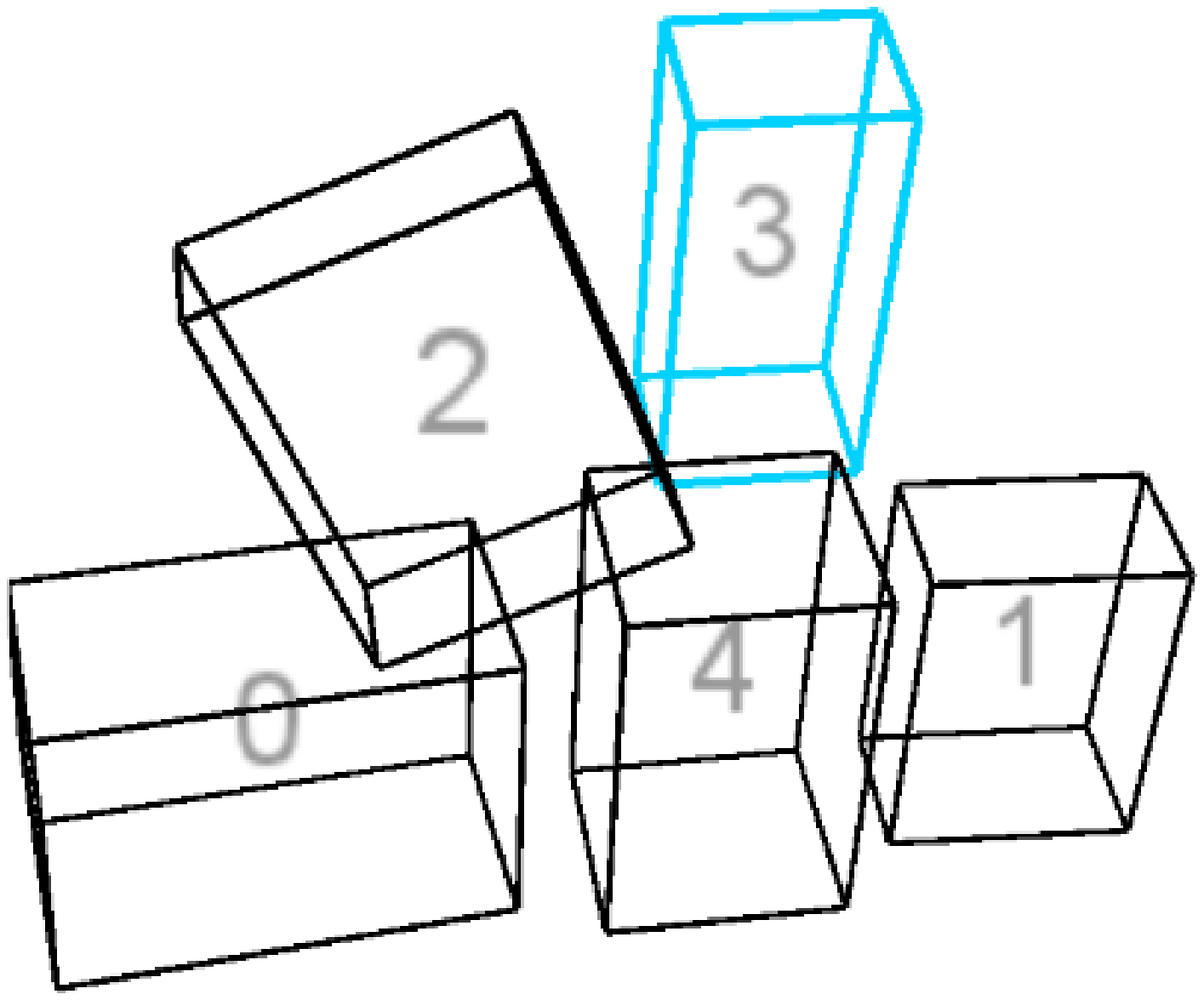}\\ \\
\includegraphics[width=\graphwidth]{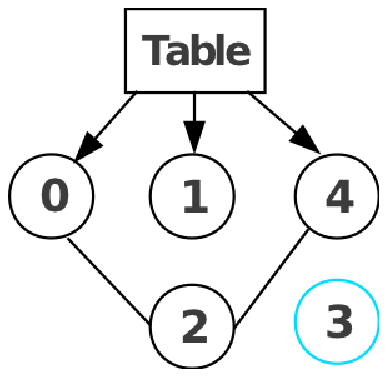}&
\begin{tabular}[b]{ll}
false(0)=0.074, hidden(0)=0.418\\
false(1)=0.080, hidden(1)=0.376\\
false(2)=0.518, hidden(2)=0.510\\
false(3)=0.106, \textcolor{cyan}{hidden(3)=0.852}\\
false(4)=0.090, hidden(4)=0.350
\end{tabular}
\\[1em]
\end{tabular}
    \caption{Experimental result 3. The input stereo image
      (upper left), estimated 6D poses (upper right), the
      resulting scene graph (lower left) and the query probability (lower right) are shown. False estimates and objects implying hidden objects are highlighted in red and cyan respectively. }
    \label{fig-p3:result3}
\end{figure*}

\begin{figure*}
\centering
\begin{tabular}{p{\figurewidth} p{\tablespacer} p{\figurewidth}}
\includegraphics[width=\figurewidth]{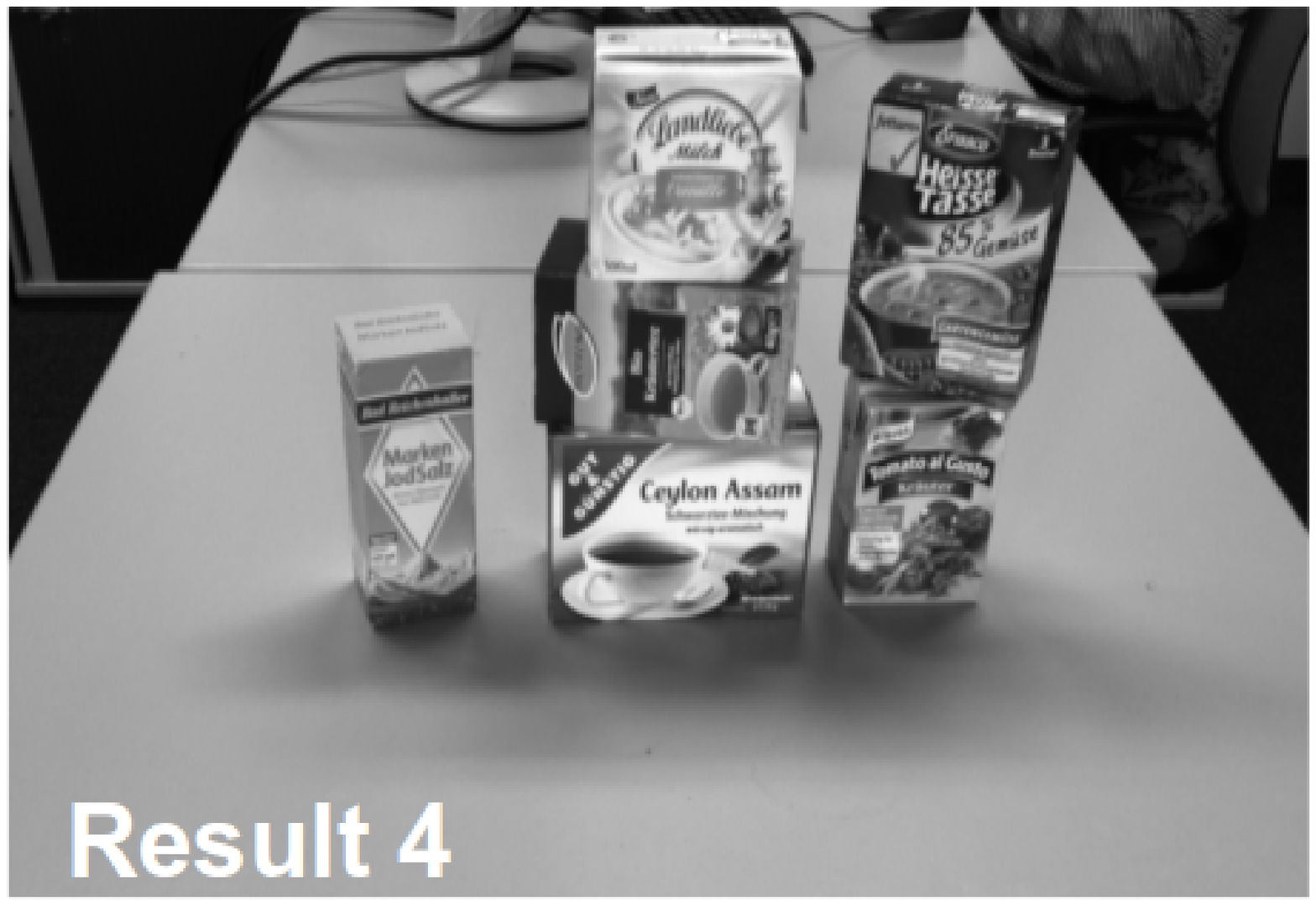} & &\includegraphics[width=\figurewidth]{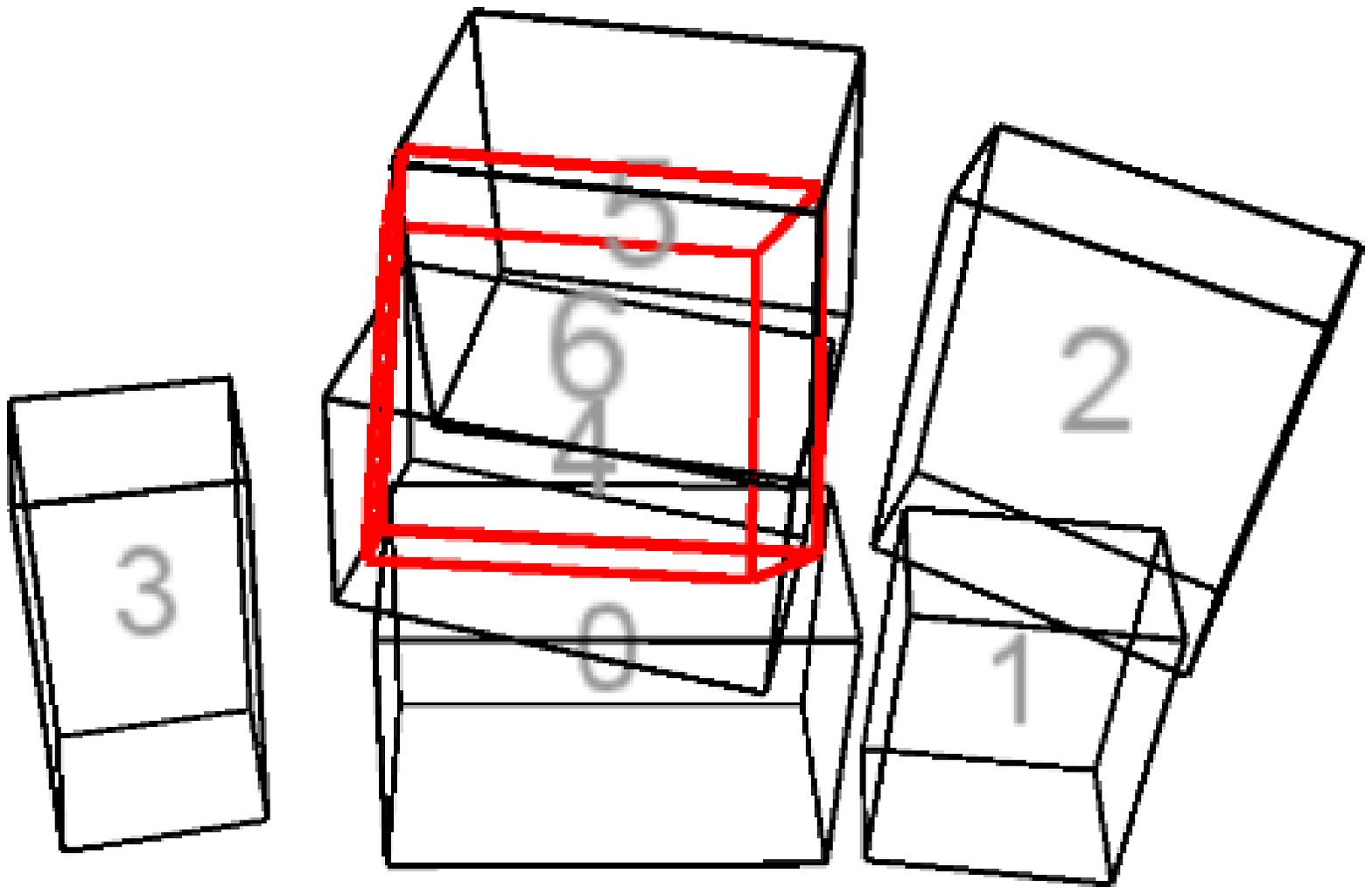}\\ \\
\includegraphics[width=\graphwidth]{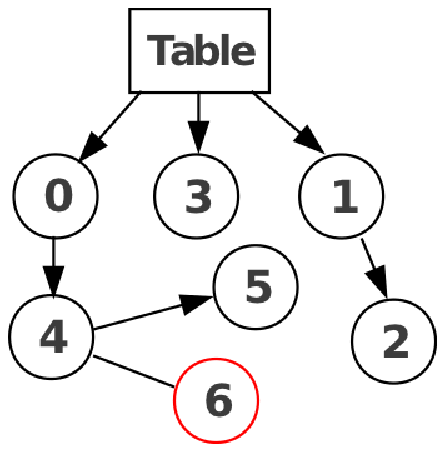}&
\begin{tabular}[b]{ll}
false(0)=0.074, hidden(0)=0.406\\
false(1)=0.036, hidden(1)=0.424\\
false(2)=0.092, hidden(2)=0.384\\
false(3)=0.102, hidden(3)=0.386\\
false(4)=0.094, hidden(4)=0.400\\
false(5)=0.130, hidden(5)=0.348\\
\textcolor{red}{false(6)=0.912}, hidden(6)=0.466
\end{tabular}
\\[1em]
\end{tabular}
    \caption{Experimental result 4. The input stereo image
      (upper left), estimated 6D poses (upper right), the
      resulting scene graph (lower left) and the query probability (lower right) are shown. False estimates and objects implying hidden objects are highlighted in red and cyan respectively. }
    \label{fig-p3:result4}
\end{figure*}

\begin{figure*}
\centering
\begin{tabular}{p{\figurewidth} p{\tablespacer} p{\figurewidth}}
\includegraphics[width=\figurewidth]{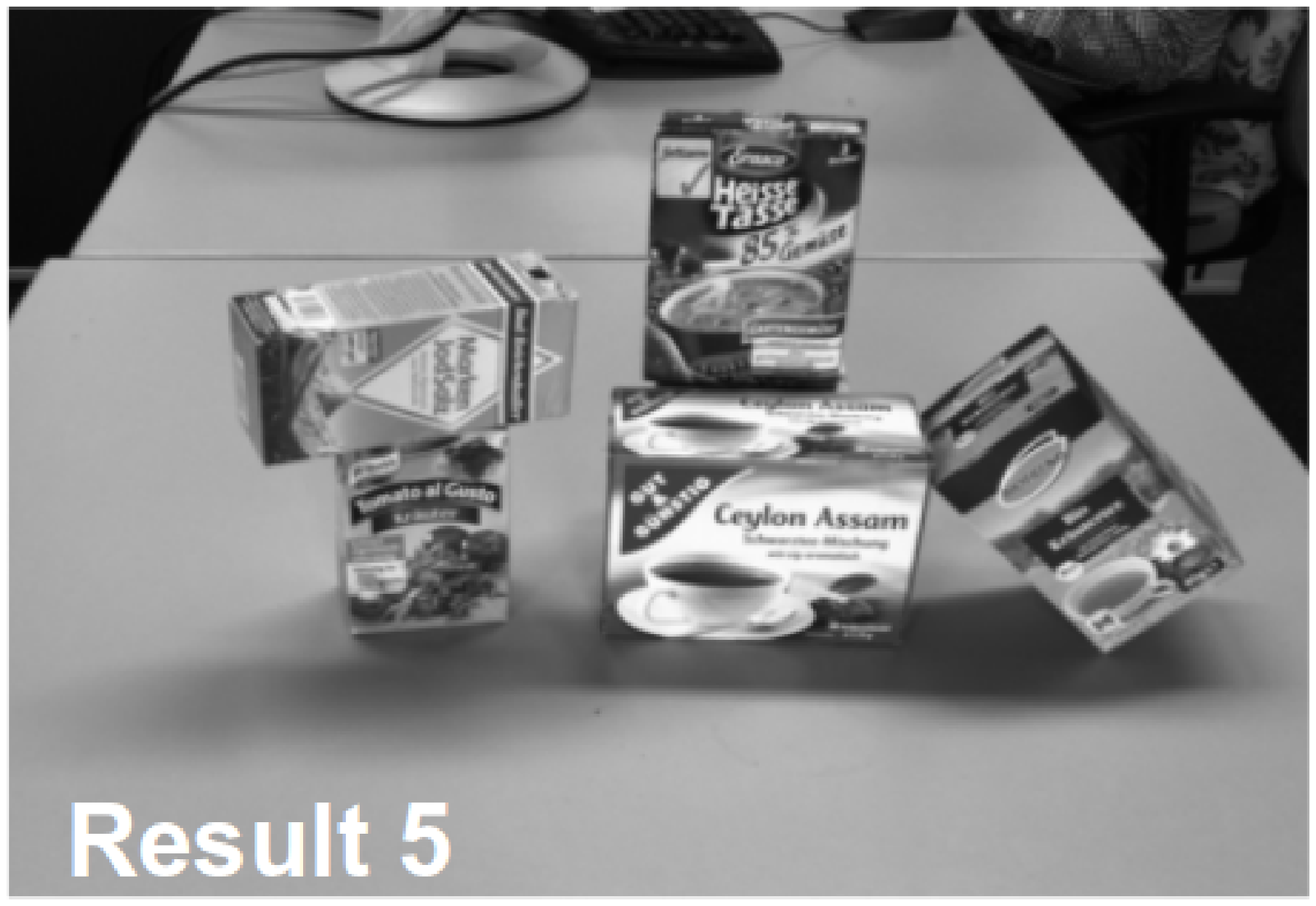} & &\includegraphics[width=\figurewidth]{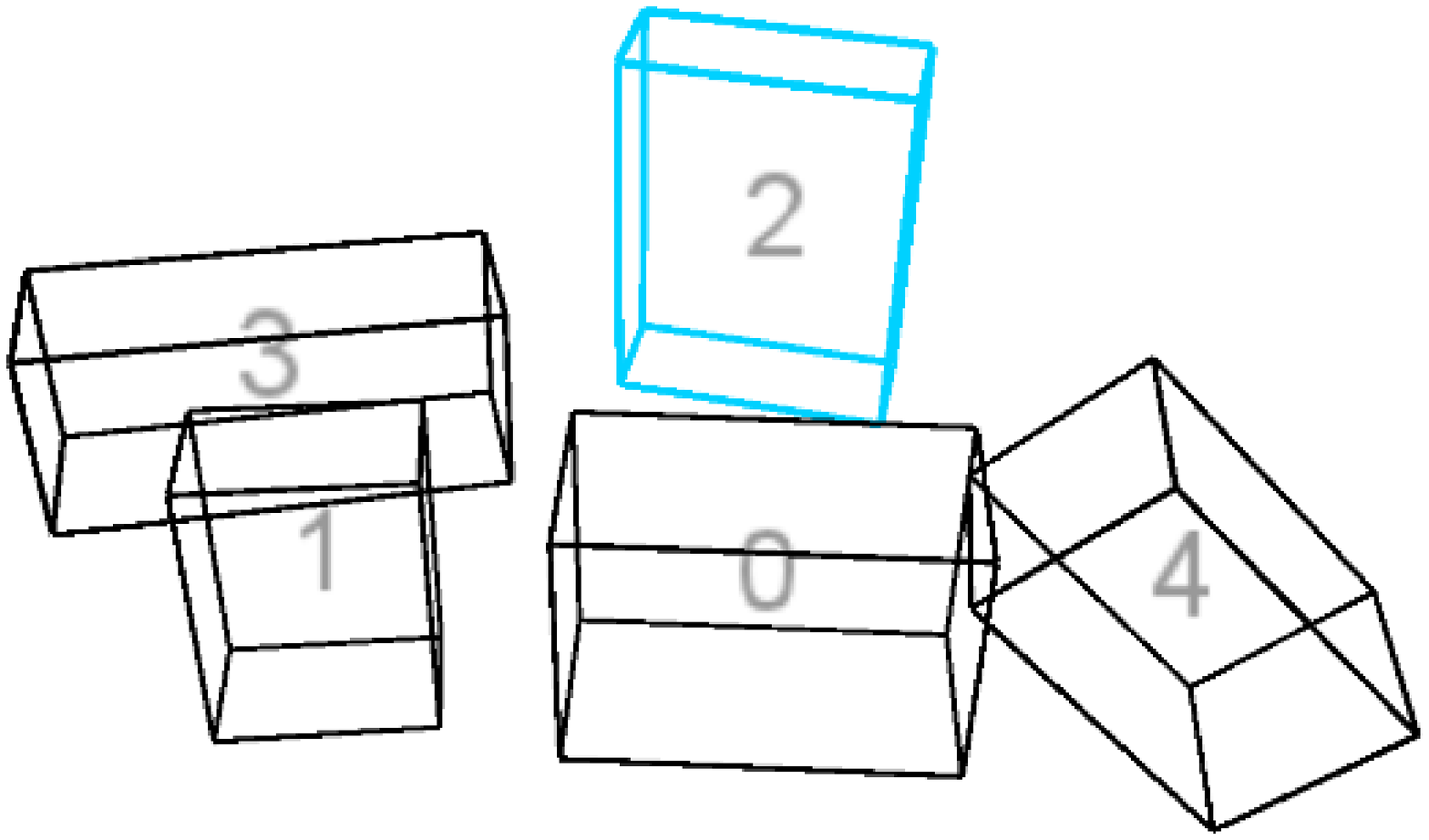}\\ \\
\includegraphics[width=\graphwidth]{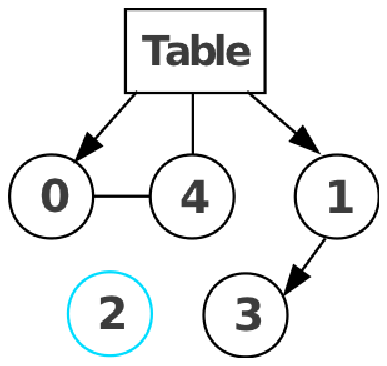}&
\begin{tabular}[b]{ll}
false(0)=0.082, hidden(0)=0.428\\
false(1)=0.044, hidden(1)=0.432\\
false(2)=0.150, \textcolor{cyan}{hidden(2)=0.816}\\
false(3)=0.110, hidden(3)=0.382\\
false(4)=0.496, hidden(4)=0.478
\end{tabular}
\\[1em]
\end{tabular}
    \caption{Experimental result 5. The input stereo image
      (upper left), estimated 6D poses (upper right), the
      resulting scene graph (lower left) and the query probability (lower right) are shown. False estimates and objects implying hidden objects are highlighted in red and cyan respectively. }
    \label{fig-p3:result5}
\end{figure*}

\begin{figure*}
\centering
\begin{tabular}{p{\figurewidth} p{\tablespacer} p{\figurewidth}}
\includegraphics[width=\figurewidth]{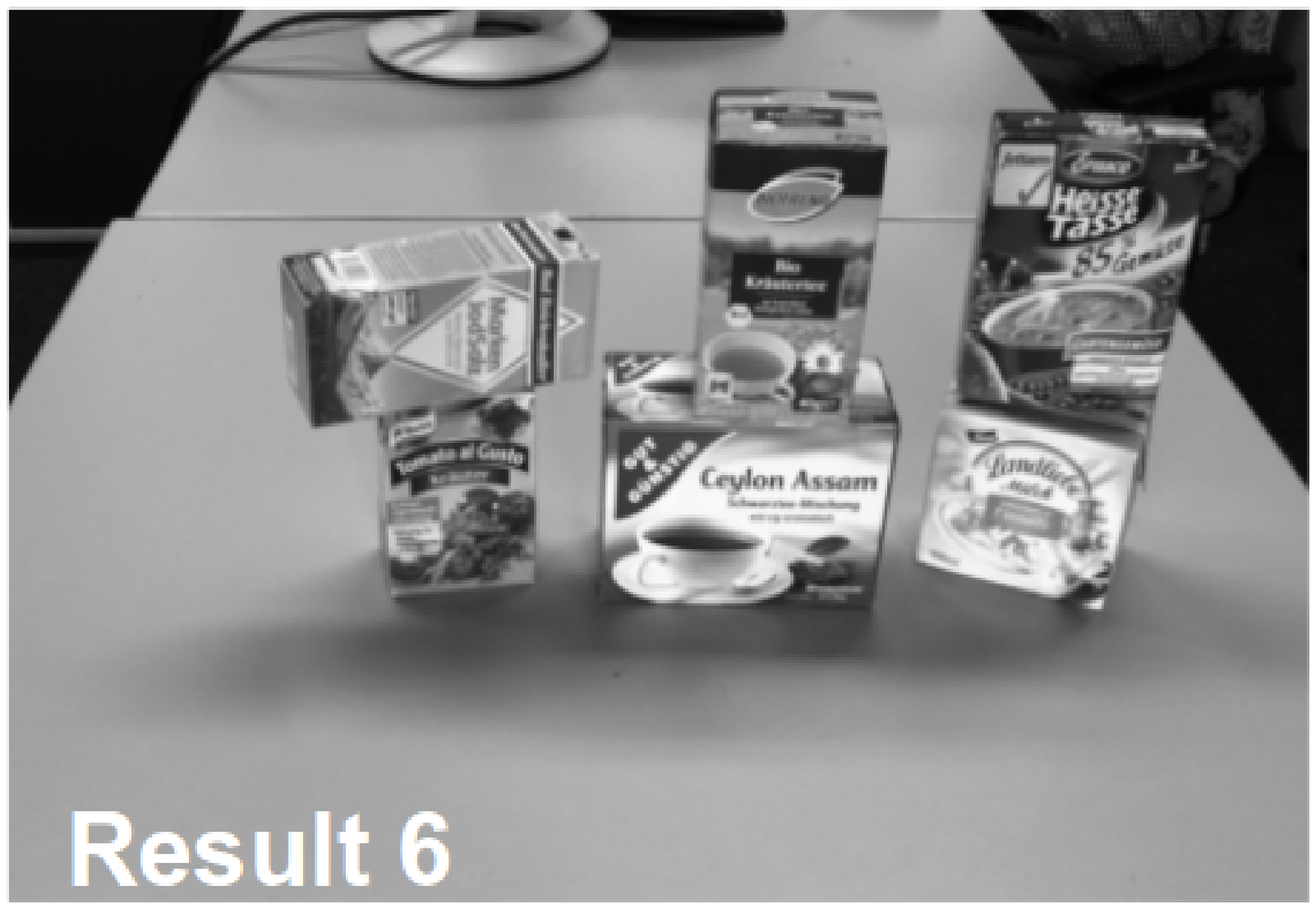} & &\includegraphics[width=\figurewidth]{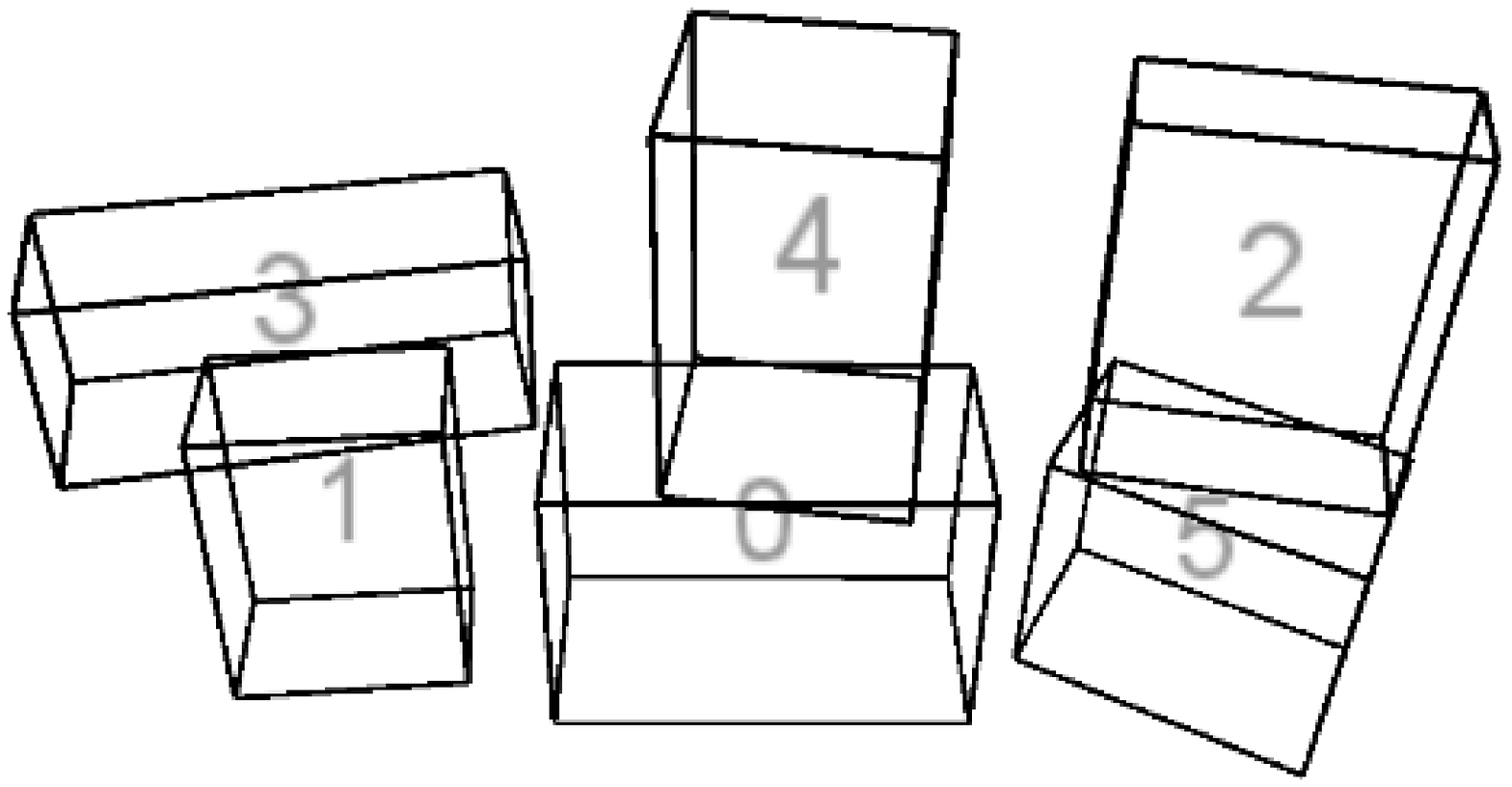}\\ \\
\includegraphics[width=\graphwidth]{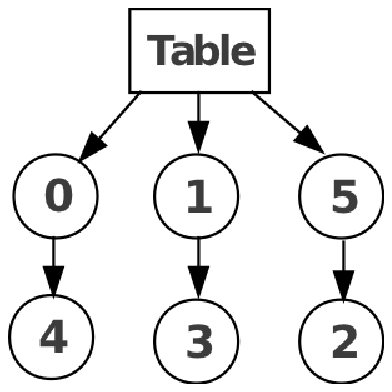}&
\begin{tabular}[b]{ll}
false(0)=0.070, hidden(0)=0.416\\
false(1)=0.036, hidden(1)=0.454\\
false(2)=0.158, hidden(2)=0.384\\
false(3)=0.110, hidden(3)=0.400\\
false(4)=0.152, hidden(4)=0.374\\
false(5)=0.074, hidden(5)=0.358
\end{tabular}
\\[1em]
\end{tabular}
    \caption{Experimental result 6. The input stereo image
      (upper left), estimated 6D poses (upper right), the
      resulting scene graph (lower left) and the query probability (lower right) are shown. False estimates and objects implying hidden objects are highlighted in red and cyan respectively. }
    \label{fig-p3:result6}
\end{figure*}

\begin{figure*}
\centering
\begin{tabular}{p{\figurewidth} p{\tablespacer} p{\figurewidth}}
\includegraphics[width=\figurewidth]{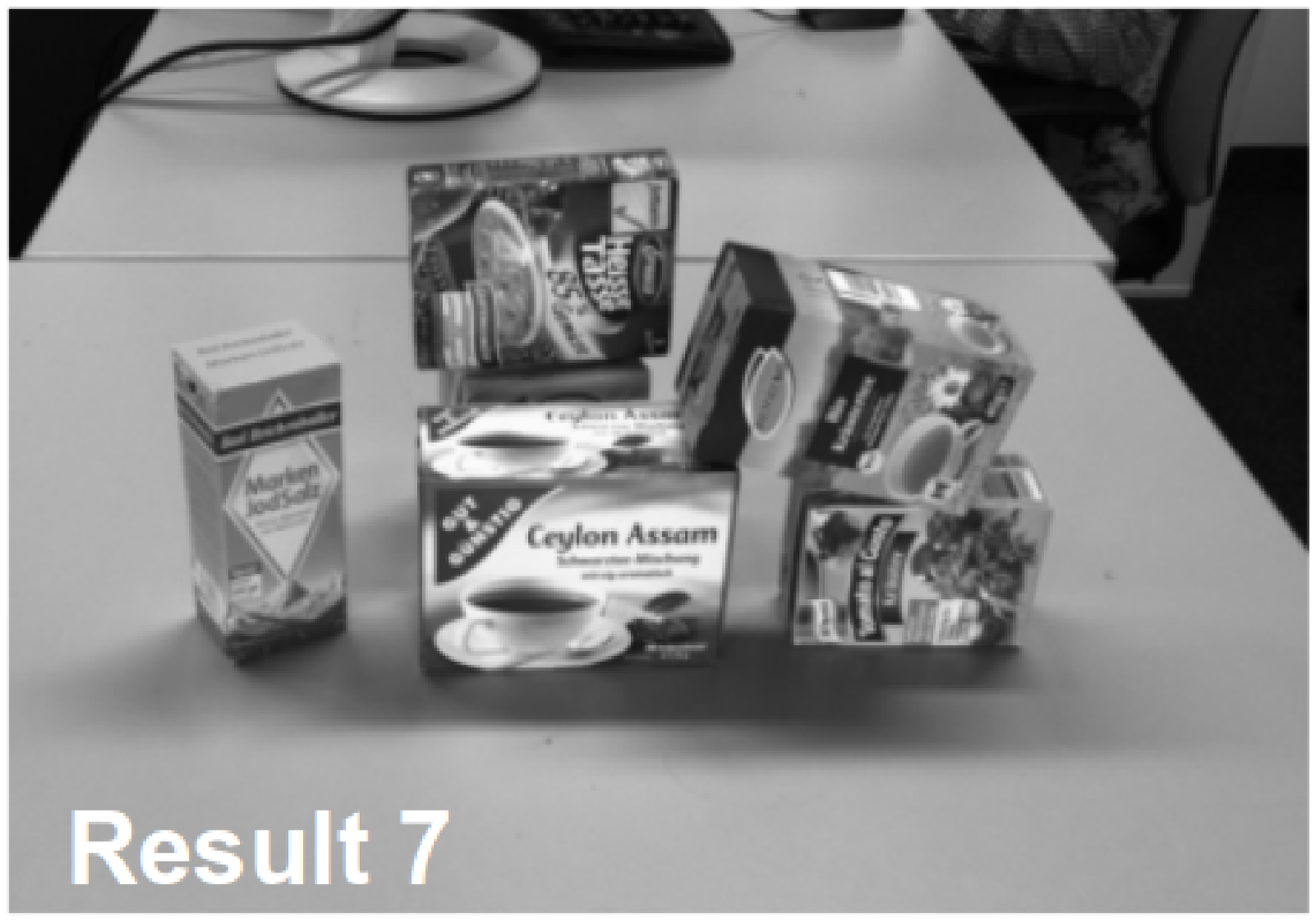} & &\includegraphics[width=\figurewidth]{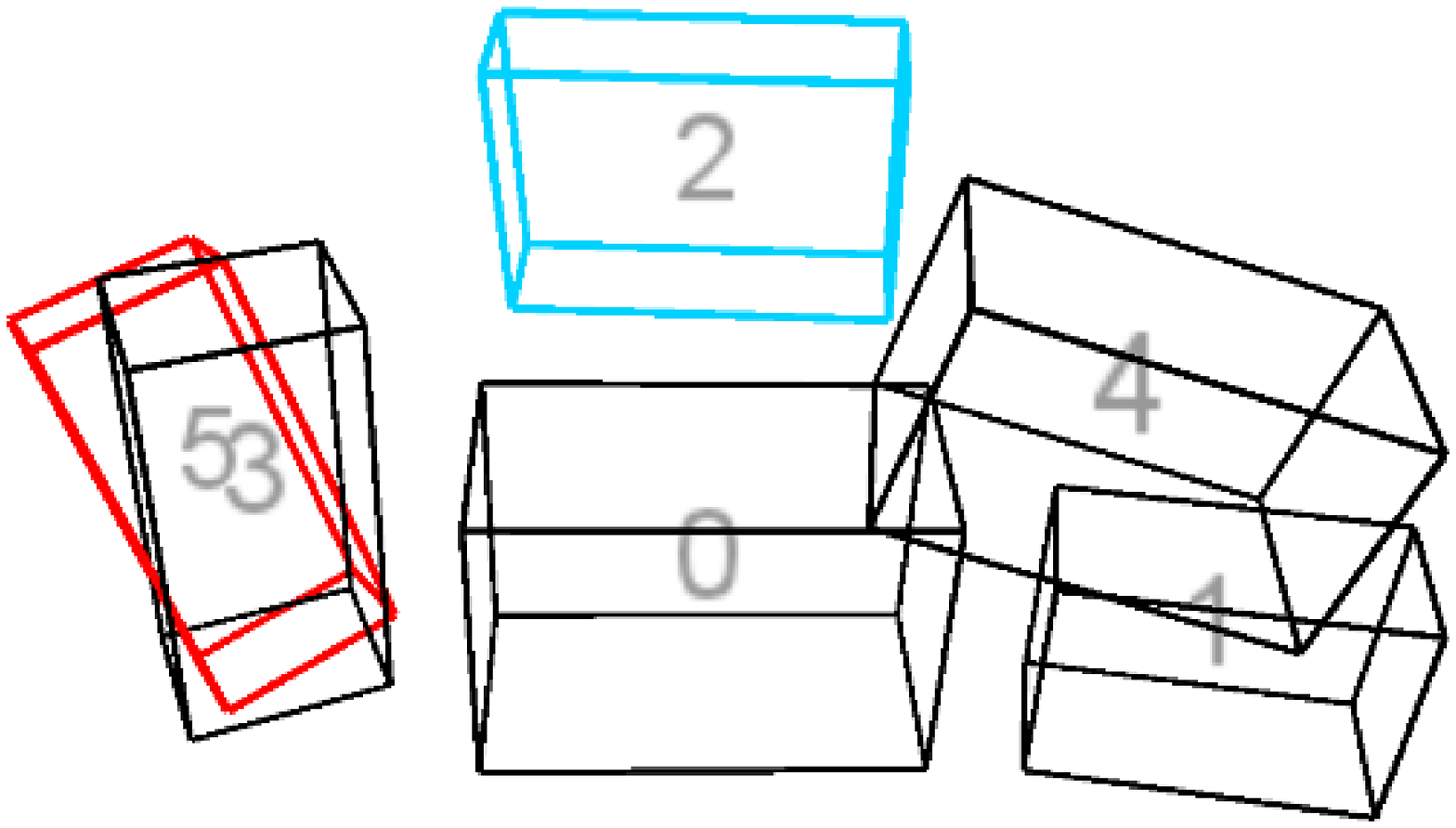}\\ \\
\includegraphics[width=\graphwidth]{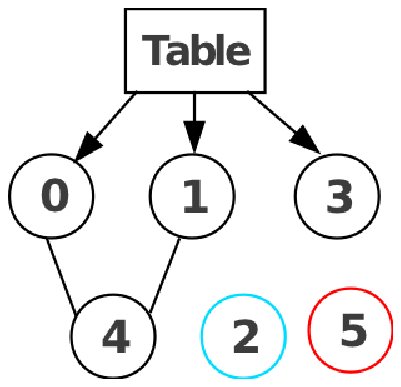}&
\begin{tabular}[b]{ll}
false(0)=0.096, hidden(0)=0.404\\
false(1)=0.088, hidden(1)=0.406\\
false(2)=0.136, \textcolor{cyan}{hidden(2)=0.808}\\
false(3)=0.138, hidden(3)=0.386\\
false(4)=0.492, hidden(4)=0.506\\
\textcolor{red}{false(5)=0.914}, hidden(5)=0.460\\
\end{tabular}
\\[1em]
\end{tabular}
    \caption{Experimental result 7. The input stereo image
      (upper left), estimated 6D poses (upper right), the
      resulting scene graph (lower left) and the query probability (lower right) are shown. False estimates and objects implying hidden objects are highlighted in red and cyan respectively. }
    \label{fig-p3:result7}
\end{figure*}

A selection of typical results is shown in Fig. \ref{fig-p3:result1} to \ref{fig-p3:result7}. In each figure, the left camera image of the stereo image, the estimated poses, the resulting scene graph and the corresponding query
probabilities are shown. False estimates and objects implying the
existence of hidden objects are highlighted in red and cyan
respectively. It can be seen, that all the perceived scenes are
correctly represented by our scene graphs. Arrows indicate that an
object stably supports another object. Undirected lines mean that two
objects have an unstable contact.

\subsection{Inference}
In our experiments, the defined knowledge base (Table \ref{TAB:formulas}) is used to reason about false estimates and hidden objects in the perceived scenes. In all experiments, the false estimates and hidden objects are correctly inferred. In the used
MLN tool \cite{probcog}, the query probabilities are calculated based
on certain sampling methods, and their values $v$ are normalized ($v
\in [0,1]$). We interpret these values as follows:
\begin{itemize}
	\item If the value is around 0.5, i.e., $0.4 < v < 0.6$, the
          uncertainty of the corresponding query is the biggest, and
          we do not make decisions, e.g., \emph{false(2)} and
          \emph{hidden(0)} in result \#3.
	\item If the value is greater than a given threshold, i.e.,
          $v > 0.6$, the corresponding query is considered to be true, e.g.,
          \emph{false(5)} and \emph{hidden(2)} in result \#7.
	\item If the value is lower than a given threshold, i.e.,
          $v < 0.4$, the corresponding query is considered to be false,
          e.g., \emph{false(0)} and \emph{hidden(4)} in result \#1.
\end{itemize}

We manually labeled 25 complex table-top scenes. Each of the scenes contained several household  objects of various types and had rather complex configurations. The 25 scenes contained in all 10 hidden objects and 5 false estimates, all of which were correctly inferred.

To check the robustness of our system, all the experiments were carried out 20 times. The generated scene graphs stay the same. In addition, false estimates and hidden objects in the scenes are also correctly inferred by the defined MLN in all repeated experiments.

\subsection{Runtime}
In experiments, we have also tested the run time performance of the proposed system.
In each iteration, the run time of our system is mainly spent on MLN
reasoning (including evidence generation) and the MCMC process. With
a single-threaded implementation on an Intel i7 CPU, the average
processing time of each iteration for the experiments shown in this
paper is 2.18 seconds. 68.8\% of this processing time is spent on MLN
reasoning, and the other 31.2\% is spent on the MCMC process. To get
a good scene graph of the perceived scene, our system needs to perform 10
to 15 iterations.

\section{Conclusion}
\label{sec:sum}
In this paper, we used our knowledge-supervised MCMC sampling technique to model table-top scenes. Our system, as a whole, demonstrates a probabilistic approach to generate
abstract scene graphs for table-top scenes using object pose
estimation as input. Our approach explicitly makes use of
task-specific context knowledge by defining this knowledge as
descriptive logic rules in Markov logic. Integrating these with
a probabilistic sensor model, we perform maximum posterior estimation of
the scene parameters using our knowledge-supervised MCMC process.

We evaluated our approach using real world scenes. Experimental
results confirm that our approach generates correct scene graphs which
represent the perceived table-top scenes well. By reasoning in the
defined MLN, false estimates of the object poses and hidden objects of
the perceived scenes were correctly inferred.

Currently, objects with a regular shape that can be well represented by an oriented bounding box are used for scene analysis. This box shape is mainly used to simplify the discriminative evidence generation. A possible future direction could be the extension to objects with irregular shapes.

\section*{Acknowledgements}

This work is accomplished with the support of the Technische Universit\"at M\"unchen - Institute for Advanced Study, funded by the German Excellence Initiative.




\bibliographystyle{elsarticle-num}
\bibliography{overall-bibliography}




%
%
%

\break

\end{document}